\documentclass[10pt,conference]{IEEEtran}
\IEEEoverridecommandlockouts
\usepackage{cite}
\usepackage{amsmath,amssymb,amsfonts,amsthm}
\usepackage{algorithmic}
\usepackage{graphicx}
\usepackage{subfigure}
\usepackage{textcomp}
\usepackage{xcolor}
\usepackage{multirow}
\usepackage[ruled,vlined]{algorithm2e}
\usepackage{framed}

\newtheorem{myDef}{Definition}
\newtheorem{myproof}{Proof}
\newtheorem{mythe}{Theorem}
\def\BibTeX{{\rm B\kern-.05em{\sc i\kern-.025em b}\kern-.08em
    T\kern-.1667em\lower.7ex\hbox{E}\kern-.125emX}}
\begin{document}

\title{Boosting the Robustness Verification of DNN by Identifying the Achilles's Heel
}

\author{\IEEEauthorblockN{Chengdong Feng$^1$, Zhenbang Chen$^1$, Weijiang Hong$^{1,2}$, Hengbiao Yu$^{1,2}$, Wei Dong$^1$, Ji Wang$^{1,2}$}
\IEEEauthorblockA{$^1$College of Computer, National University of Defense Technology, Changsha, China\\
$^2$State Key Laboratory of High Performance Computing, National University of Defense Technology, Changsha, China\\
fengchendong@outlook.com, zbchen@nudt.edu.cn, weijiang\_h@163.com, \{hengbiaoyu, wdong, wj\}@nudt.edu.cn}
}
%, 

%\and
%\and
%\and
%\and

\maketitle

\begin{abstract}
Deep Neural Network (DNN) is a widely used deep learning technique. How to ensure the safety of DNN-based system is a critical problem for the research and application of DNN. Robustness is an important safety property of DNN. However, existing work of verifying DNN's robustness is time-consuming and hard to scale to large-scale DNNs. %Scalability is the key challenge for the verification of DNN. 
In this paper, we propose a boosting method for DNN robustness verification, aiming to find counter-examples earlier. Our observation is DNN's different inputs have different possibilities of existing counter-examples around them, and the input with a small difference between the largest output value and the second largest output value tends to be the achilles's heel of the DNN. We have implemented our method and applied it on Reluplex, a state-of-the-art DNN verification tool, and four DNN attacking methods. The results of the extensive experiments on two benchmarks indicate the effectiveness of our boosting method.
%$\\$
%$\\$
%$\\$
%$\\$
%$\\$
%$\\$
%$\\$
%$\\$
%$\\$
%$\\$
%$\\$
%$\\$
%$\\$
\end{abstract}


%!TEX root = main.tex
\section{Introduction}

Nowadays, the research and application of deep learning (DL) \cite{daglib} have achieved tremendous progresses. Deep learning has been widely used in many areas, including speech recognition \cite{X12a}\cite{MikolovDPBC11}, autonomous driving \cite{KimP17}, image classification \cite{KrizhevskySH12}, \emph{etc}. Deep learning techniques, \emph{e.g.}, Deep Neural Network (DNN) \cite{Schmidhuber15}, play a crucial role in the products or systems in these areas.  When applied in critical areas, such as autonomous cars \cite{KimP17} and airborne collision avoidance systems \cite{Julian2016Policy}, DL-based systems need to assure high-quality safety. However, there already exist the cases in which DL-based systems cause disasters, such as the one caused by Tesla car in 2016 \cite{tesla}. Besides, there exist studies \cite{DeepXplore}\cite{DeepTest} that use adversarial examples \cite{SzegedyZSBEGF13} to attack DL-based systems. How to ensure the safety of DL-based systems is challenging.

% Existing work

DNN is a representative DL classification technique. The existing work of safety assurance for DNN-based systems mainly focuses on the \emph{robustness} of DNN. Existing methods have two categories: 1) attacking methods \cite{GoodfellowSS14}\cite{KurakinGB16}\cite{PapernotMJFCS16}\cite{Moosavi-Dezfooli16}\cite{Carlini017} that generate the adversarial examples  \cite{SzegedyZSBEGF13} of a DNN and retrain the DNN to improve the robustness; 2) defense methods that include verifying the robustness of a DNN \cite{KatzBDJK17}, detecting the attacking adversarial inputs online \cite{abs-1805-05010}, \emph{etc}. 

% Problem

To verify the robustness of a DNN, the existing work usually models the DNN, such as symbolic encoding \cite{abs-1807-10439} and abstraction \cite{GehrMDTCV18}, and tries to verify the model via symbolic solving or invariant checking. If the verification succeeds, the DNN is \emph{proved} to be robust; otherwise, \emph{counter-examples} (or adversarial examples) are produced. The existing verification methods differ in the aspects including scalability, completeness, \emph{etc}. However, verification's cost is usually high, and the verification methods are difficult to scale to large DNNs. How to improve the scalability  is a key problem for verifying DNNs.

%Our observation and basic idea

Existing studies \cite{GoodfellowSS14}\cite{KurakinGB16}\cite{PapernotMJFCS16}\cite{Moosavi-Dezfooli16}\cite{Carlini017} of adversarial examples \cite{SzegedyZSBEGF13} indicate that most real-world DNNs tend to have adversarial examples, \emph{i.e.}, they are not robust. 
%Existing verification methods are only feasible for verifying \emph{local robustness}, \emph{i.e.}.  
Hence, the verification methods usually produce counter-examples. Then, boosting counter-example finding during the process of verification directly improves the scalability. Most existing DNN verification methods support only the verification of \emph{local robustness}, \emph{i.e.}, given an initial point $p$ of the input domain, prove that the neighbouring points within a limited range $\delta$ are classified into the same type as $p$. Hence, the selections of $p$ and $\delta$ directly influence the result and the efficiency of verification. We observe that verification tools tend to quickly find counter-examples at some initial points. Therefore, if we can select the right points, we can boost finding counter-examples.

In this paper, we first prove that the outputs of a DNN using ReLU are continuous \emph{w.r.t.} the inputs. Then, based on the continuity result, we propose a method for evaluating the possibility of finding counter-examples around an input. The key idea is an input whose largest and second largest outputs are close is likely to have counter-examples around. As far as we know, it is the first evaluation method considering this aspect. Besides, we propose a lightweight pre-analysis to boost finding counter-examples further. We have implemented our boosting method and applied it on Reluplex \cite{KatzBDJK17}, \emph{i.e.}, a state-of-the-art robustness verification tool for DNN, and representative DNN attacking methods. The experimental results on two benchmarks, \emph{i.e.}, ACAS-Xu \cite{Julian2016Policy} and MNIST \cite{MNIST}, indicate the effectiveness of our method. 

%Our results

The main contributions of this paper are as follows:
\begin{itemize}
%\item The continuity property proof of the DNNs using ReLU.
\item Based the continuity property of DNN, we propose an evaluation method for selecting the inputs around which counter-examples tend to exist.
\item We propose a pre-analysis greedy algorithm to speed up counter-example finding further.
\item We have implemented our boosting method and applied it on Reluplex and four adversarial example generation methods. The extensive experiments on two representative DNN benchmarks indicate: compared with random method, our boosting method can achieve at least an order of magnitude time speedup in finding the same amount of counter-examples; under the same time budget, our method can find an order of magnitude more counter-examples; besides, our method can averagely improve the success rate of the methods for generating adversarial examples by 3.2 times. 
\end{itemize}

The remaining of this paper is organized as follows. \mbox{Section 2} briefly introduces the backgrounds and motivations of our method. Section 3 proves the continuity of the DNNs using ReLU. Section 4 presents our boosting method. Section 5 gives experimental results. Section 6 reviews the related work, and the conclusion is drawn in Section 7.

%!TEX root = main.tex
\section{Preliminary and Motivation}
\label{sec:P&F}
In this section, we will briefly introduce the basic concepts of DNN, its robustness, and the DNN verification tool Reluplex. Then, our boosting method will be motivated. 

\subsection{DNN}
\label{subsec:dnn}
Generally, a DNN is comprised of multiple layers, wherein each layer consists of nodes called \emph{neurons}. The first and the last layer are \emph{input} layer and \emph{output} layer, respectively. The remaining layers are \emph{hidden} layers. Here, we focus on the feed-forward multi-layer neural networks\cite{Schmidhuber15}. For each neuron of every layer, it is connected to each neuron of the next layer in a forward direction with a \emph{weight}. Each neuron in the \emph{hidden} layers has an incoming value and an outgoing value, which are linked by an \emph{activation function}, such as \emph{ReLU}\cite{NairH10}, \emph{tanh}\cite{Bailey2006Tanh} and \emph{sigmoid}\cite{HanM95}. In this paper, we only consider the DNN using ReLU activation function, \emph{i.e.}, $\mathit{max}(0, x)$. If a DNN has $n$ hidden layers and the $i$th layer has $m$ neurons, we use $n_{i,j}$ to denote the $j$th neuron of the $i$th layer, where $i\in\{0,...,n+1\}$ and $j\in\{0,...,m-1\}$. Then, we use $v_{i,j}$ to denote the outgoing value of $n_{i,j}$, and $w_{i,j}^{k}$ to denote the weight between $n_{i,j}$ and $n_{i+1,k}$.

Given an input from the \emph{input} layer, the DNN propagates the values of this input from layer to layer \emph{w.r.t.} the weights and the activation function. The outgoing values of the output layer are the results of the DNN.  If the DNN is used for classification, then a neuron in the output layer represents a category. The output layer's neuron with the largest outgoing value is the classification result. Here, we focus on the DNNs for classification task. That is, given an  input $x_0$, a well-trained DNN $N$ and a set of category labels $\{c_1, c_2, ... , c_n\}$, the result label of $x_0$ is denoted as $N(x_0)$, which belongs to $\{c_1, c_2, ... , c_n\}$. Besides, given an input $x$, we use $N_{d}(x)$ to represent $V_1(N, x) - V_2(N, x)$, where $V_1(N, x)$ and $V_2(N, x)$ are $N$'s largest and second largest outgoing values in the output layer when calculating the input $x$, respectively.

%\begin{equation} 
%R(x)=\left\{
%\begin{aligned}
%x  & & x > 0 \\
%0  & & otherwise
%\end{aligned}
%\right.
%\end{equation}

\begin{figure}[!ht]
  \centering
  \includegraphics[width=0.35\textwidth]{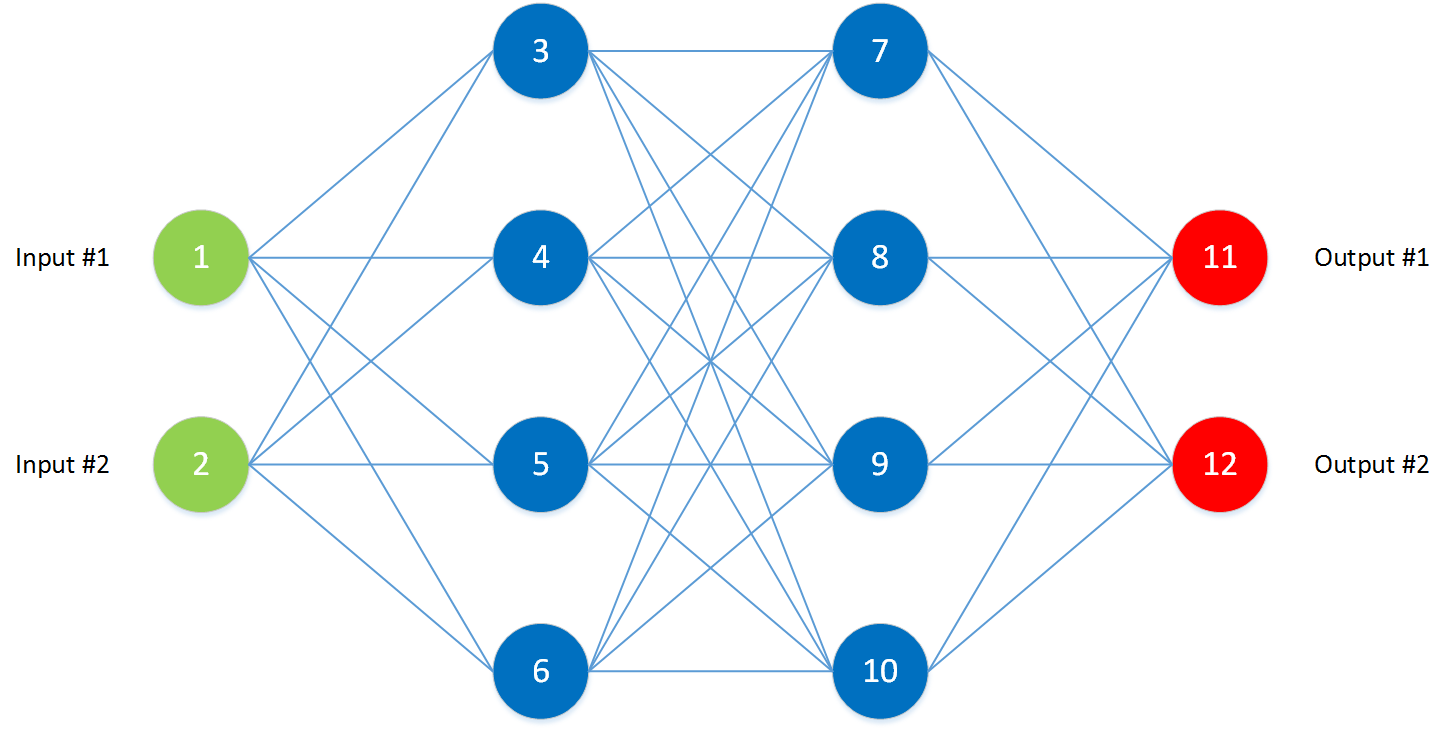} 
  \caption{An Example DNN.} 
  \label{DNN} 
\end{figure}

As an example, Figure \ref{DNN} illustrates a simple DNN with only 2 hidden layers. Each layer has four neurons. Both of input and output layers have two neurons. Then, for example, the outgoing value $v_{1, 0}$ of the 3rd neuron $n_{1,0}$ is calculated as follows:
\begin{equation}
v_{1, 0} := max(0, v_{0,0}*w_{0,0}^{0} + v_{0,1}*w_{0,1}^{0})
\end{equation}
For the 11th neuron $n_{3,0}$ in the output layer, its outgoing value $v_{3, 0}$ is calculated as follows:
\begin{equation}
v_{3, 0} := v_{2,0}*w_{2,0}^{0} + v_{2,1}*w_{2,1}^{0} + v_{2,2}*w_{2,2}^{0} + v_{2,3}*w_{2,3}^{0}
\end{equation}

\subsection{Robustness of DNN}
Robustness is an important safety property of DNN. Especially, the robustness we consider here is referred to the adversarial robustness \cite{abs-1709-02802}. That is, given a well-trained DNN and a correctly classified input $x$, if we modify this input by adding some small imperceptible perturbations, then this input should be classified to the same category as $x$. Robustness can be divided into two categories: \emph{global adversarial robustness} and \emph{local adversarial robustness}. In the following, we give their formal definitions. 
\begin{myDef}
{\bf{Global adversarial robustness: }} Given a DNN network $N$ and a distance $\delta$, $N$ is global adversarial robustness iff for any two points $x_1$ and $x_2$ in the input space that satisfy $\vert\vert x_1 - x_2\vert\vert \le \delta$, then we require that $N(x_1) = N(x_2)$.
\end{myDef}

However, because most DNNs have adversarial examples \cite{SzegedyZSBEGF13}, the global adversarial robustness is almost impossible. A weaker version of adversarial robustness, \emph{i.e.}, \emph{local adversarial robustness}, is investigated by the existing DNN verification approaches. The definition of local robustness is as follows.
%As we can see, to guarantee the global adversarial robustness is almost out of the question, so we focus on a weaker adversarial robustness, \emph{i.e.}, \emph{local adversarial robustness}.

\begin{myDef}\label{local-rb-def}
{\bf{Local adversarial robustness: }}  Given a DNN network $N$, a distance $\delta$ and an initial point $x_0$, $N$ is local adversarial robustness w.r.t. $x_0$ iff for each point $x$ in the input space that satisfies $\vert\vert  x - x_0\vert\vert \le \delta$,  $N(x) = N(x_0)$.
\end{myDef}

\noindent
Hence, local robustness ensures the safety of the points around the initial point. In this paper, we focus on verifying local robustness.

%\subsection{Reluplex}
%\label{reluplex}
Reluplex \cite{KatzBDJK17} is an SMT-based verification tool for DNN. Reluplex supports the verification of safety properties, including local robustness, global robustness, \emph{etc}. Reluplex symbolically encodes the DNN under verification and the properties to verify. Then, Reluplex transforms the verification problem into an SMT solving problem \cite{BarrettSST09}. The authors of Reluplex design a highly efficient simplex-based algorithm for solving the verification problem of the DNNs using ReLU. Reluplex shows a great effectiveness for verifying the ACAS-Xu DNNs \cite{Julian2016Policy}. If we use Reluplex to verify the local robustness of a DNN, the property to verify by Reluplex is the \emph{negation} of the local robustness. Hence, Reluplex will report \verb"UNSAT" if the DNN satisfies the local robustness, and \verb"SAT" if the DNN does not. When Reluplex reports \verb"SAT", \emph{i.e.}, an adversarial example exists, Reluplex produces a counter-example input.

\subsection{Motivation}

We want to boost DNN verification tool's procedure of finding counter-examples. To verify the local robustness of a DNN, we need to provide an initial input  to the verification tool. In principle, the initial input selection directly influences the verification result. Hence, if we can choose the inputs around which adversarial examples are likely to exist, we can directly boost the counter-example finding of verification.

%theoretical analysis
According to the definitions of DNN, the relation between DNN's input and output can be represented as a function. If ReLU activation function is used by the DNN, we observe that the DNN's function is continuous, whose proof  will be presented in Section \ref{proof}. Then, according to robustness definitions, for an input $x$, if $x$'s largest outgoing value of the output layer is very close to the second largest outgoing value, $x$ tends to be not locally robust, \emph{i.e.}, small perturbations may change the classification label of $x$. For example, for the DNN in Figure \ref{DNN}, if $v_{3, 0}$ and $v_{3, 1}$ of an input $x_0$ are very close, suppose $v_{3, 0} > v_{3, 1}$, \emph{i.e.}, $N_d(x_0)$ is small, we consider $x_0$ as the input around which adversarial examples are likely to exist, because a small change to $x_0$ may result in $v_{3, 1} > v_{3, 0}$. On the contrary, if the largest outgoing value is much bigger than the second largest outgoing value, $x$ tends to be robust, because a small change of $x$ does not change the relation of output values. 

\begin{figure}[!ht]
  \centering 
  \includegraphics[width=0.4\textwidth]{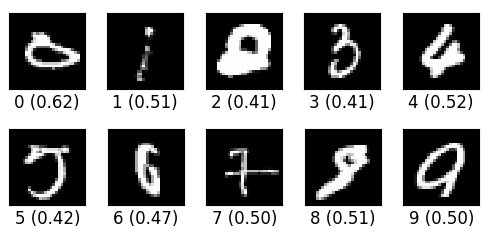}
  \caption{MNIST images with smallest difference value.}
  \label{MNIST_example}
\end{figure}

%empirical analysis
We validate our intuition on MNIST benchmark for handwritten digit recognition. The trained DNN's test accuracy is 98.02\%. Figure \ref{MNIST_example} shows the images with the smallest  difference between the largest value and the second largest value of the output layer. As shown by the figure, the images are not clear and hard to recognize, even for human. Hence, these images are likely to have counter-examples around. 
\vspace{-2mm}
\vspace{-2mm}

\begin{figure}[!ht]
  \centering
%  \subfigure[Original result]{
%    \includegraphics[width=0.45\textwidth]{Hypothesis_evaluation_origin.png}
%    \label{original}
%  }
%  \quad
%  \subfigure[Adjusted result]{
    \includegraphics[width=0.45\textwidth]{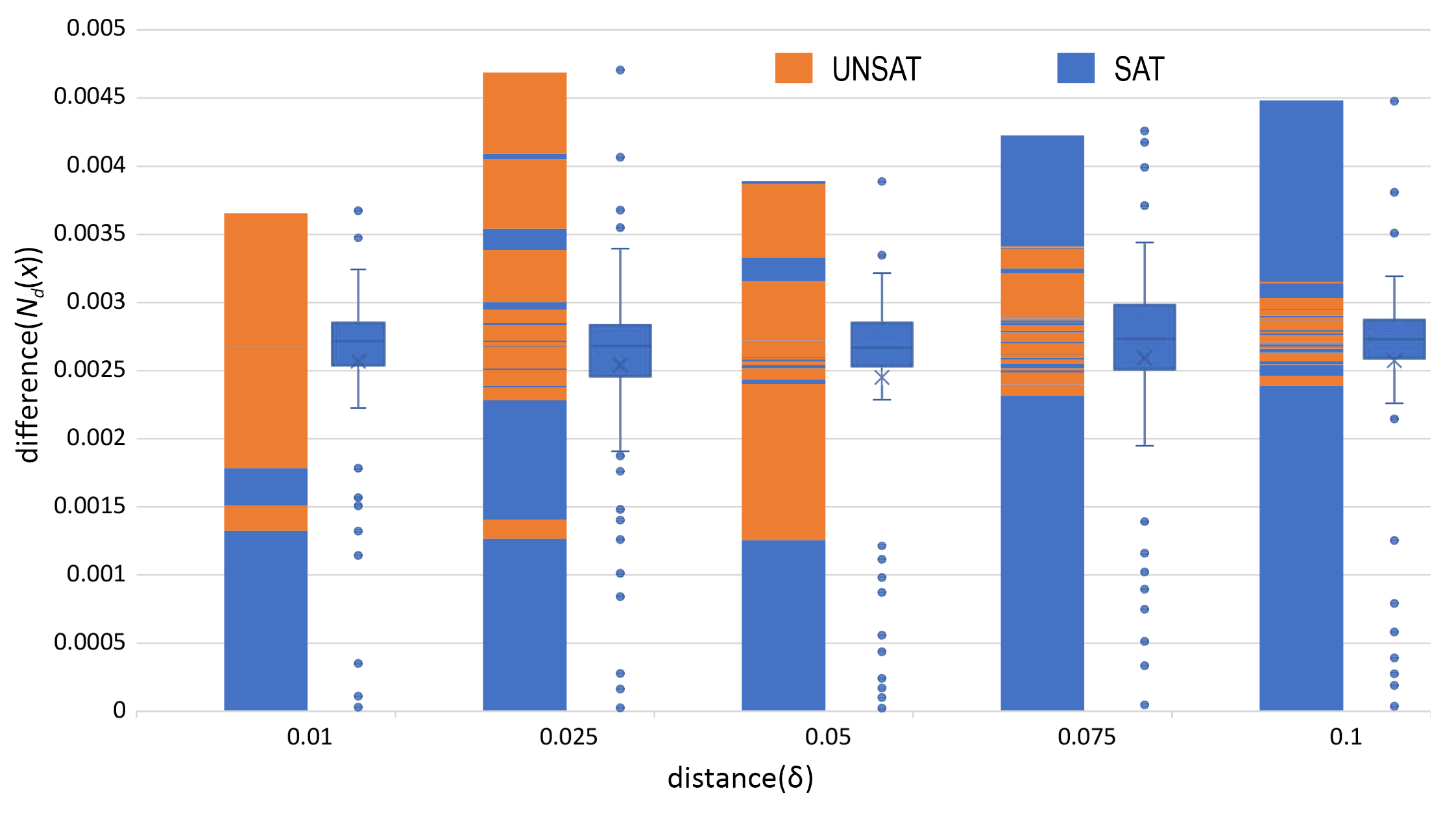}
%    \label{corrected}
%  }
\vspace{-2mm}
  \caption{The verification results under different distances.}
  \label{intuition}
\end{figure}

Besides, we have also carried out empirical studies on Reluplex. We randomly select a DNN from Reluplex's benchmark, \emph{i.e.}, ACAS-Xu \cite{Julian2016Policy}. We use Reluplex to verify the local robustness of randomly generated 100 inputs. Because distance $\delta$ is also a key parameter for verification, we carried out the experiments under 5 distances. %Figure \ref{original} shows the original result of the verification of random selection in the case of five $\delta$. For each $\delta$, we randomly selected 100 inputs for verification and plotted according to their evaluation values and results. 
Figure \ref{intuition} shows the results. The X-axis displays the values of different $\delta$ distances, and Y-axis displays the values of $N_d(x)$. 
Rectangle shows the distribution of the verification results (\verb"UNSAT" or \verb"SAT") under each distance. Box plot shows the distribution of the difference values of the inputs, and the exception points ($2\%$ to $4\%$) are removed for the sake of clarity.

%Since there is a part of the evaluation value that is much larger than most of the evaluation values, the distribution of the evaluation values is not uniform, so we have a screening of the original results to make the chart more intuitive. We removed the input with a too large evaluation value and got \textcolor{red}{Figure \ref{corrected}}, in this figure we can see that under any kind of $\delta$, the probability of SAT is greater when the evaluation value is smaller. Discarded input accounts for 2\% to 4\% of all inputs.

As shown in Figure \ref{intuition}, under each distance value, Reluplex tends to report \verb"SAT", \emph{i.e.}, finding a counter-example, when the difference value is small, which validates our intuition. Besides, during experiments, we also observe that the \verb"UNSAT"  verification time is empirically much longer than that of \verb"SAT". Hence, under the same time-budget, if we want to find more counter-examples, we need to start with the inputs around which it is more possible to have counter-examples. Furthermore, even for \verb"SAT" cases, Reluplex's verification costs are high, ranging from seconds to hours. Hence, we can consider to employ a pre-analysis (\emph{c.f.}, Section \ref{pre-analysis}) to search counter-examples before verification. If the search succeeds, we can report the counter-example instead of continuing verification; otherwise, we can still do the verification.

%!TEX root = main.tex
\section{Continuity of DNN}\label{proof}
In this section, we define and prove the continuity of DNN. In mathematics, a function $f$ is  continuous if any small change of $f$'s output can be resulted by a change of $f$'s input. The formal definition is as follows.

%Intuitively, we guess the Therefore, we wanna explore the DNN's continuity property as the DNN is also a function. At first, we give the definition of the continuous function in mathematics.

\begin{myDef}
{\bf{The continuity of function: }}The function $f : D_1 \rightarrow D_2$ is continuous iff for each input $x_0 \in D_1$, we require that for every $\epsilon > 0$ there exists a $\delta > 0$ such that for all $x \in D_1$: $\vert\vert x-x_0 \vert\vert < \delta \Rightarrow \vert\vert f(x)-f(x_0) \vert\vert < \epsilon$.
\end{myDef}

In principle, a DNN can be abstracted as a function that reveals the relation between the inputs and the outputs of the training examples. Intuitively, we guess the function of a DNN using ReLU is continuous. The basic idea of proving the continuity of DNN using ReLU is to prove the continuity of basic function first, and then prove the continuity is closed under composition operators.  

%Take the activation function \emph{ReLU} as an example. We wanna prove the following theorem.
\begin{figure*}
  \centering 
  \includegraphics[width=0.9\textwidth]{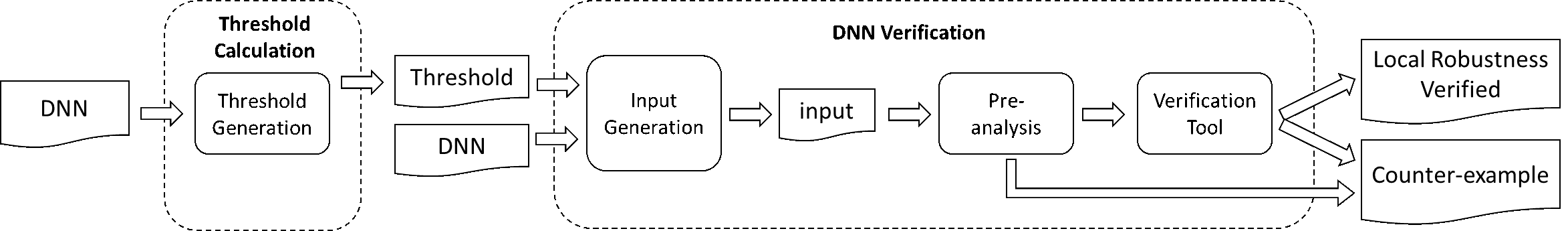}
  \caption{The framework of boosting method.}
  \vspace{-4mm}
  \label{framework}
\end{figure*}

\begin{mythe}
The activation function ReLU, i.e., $max(0,x)$, is a continuous function.
\end{mythe}

\begin{myproof}
Given a randomly chosen input $x_0$ from the input domain $D$, for every $\epsilon > 0$ and $x \in D$ that satisfies $\vert\vert x-x_0 \vert\vert < \epsilon$ (supposed that $\delta = \epsilon$), we want to prove $\vert\vert ReLU(x)-ReLU(x_0) \vert\vert < \epsilon$. According to the function itself, we split the proof goal into four cases.

Case.1 if $x < 0$ and $x_0 < 0$, then $\vert\vert ReLU(x)-ReLU(x_0) \vert\vert = 0 < \epsilon$.

Case.2 if $x \ge 0$ and $x_0 \ge 0$, then $\vert\vert ReLU(x)-ReLU(x_0) \vert\vert = \vert\vert x-x_0 \vert\vert < \epsilon$.

Case.3 if $x < 0$ and $x_0 \ge 0$, then $\vert\vert ReLU(x)-ReLU(x_0) \vert\vert = \vert\vert x_0 \vert\vert < \vert\vert x_0-x \vert\vert < \epsilon$.

Case.4 if $x \ge 0$ and $x_0 < 0$, then $\vert\vert ReLU(x)-ReLU(x_0) \vert\vert = \vert\vert x_0 \vert\vert < \vert\vert x-x_0 \vert\vert < \epsilon$.

In conclusion, theorem holds.
\end{myproof}

Besides, based on continuity's definition, we can prove that continuity property is closed under  composition operators.

\begin{mythe}\label{th2}
{\bf{The continuity of function composition: }}Given a constant $c$, two continuous functions $f$ and $g$, we have the following composition function: $c * f$, $f+g$ and $f\circ g$.  All of them are continuous. 
\end{mythe}

\begin{myproof} Cases:

{\bf{$c * f$ is continuous: }}
As $f$ is continuous, we have the following that for each input $x_0 \in D_1$ and  every $\epsilon/c > 0$ there exists a $\delta > 0$ such that for all $x \in D_1$: $\vert\vert x-x_0 \vert\vert < \delta \Rightarrow \vert\vert f(x)-f(x_0) \vert\vert < \epsilon /c$. Furthermore, we have $\vert\vert c * f(x)- c * f(x_0) \vert\vert < c * (\epsilon /c) = \epsilon$, which implies $c * f$ is continuous.

{\bf{$f+g$ is continuous: }}
As $f$ and $g$ are continuous, we have the following that for each input $x_0 \in D_1$ and every $\epsilon/2 > 0$ there exist $\delta_1 > 0$ and $\delta_2 > 0$ such that for all $x \in D_1$: $\vert\vert x-x_0 \vert\vert < \delta_1 \Rightarrow \vert\vert f(x)-f(x_0) \vert\vert < \epsilon /2$ and $\vert\vert x-x_0 \vert\vert < \delta_2 \Rightarrow \vert\vert g(x)-g(x_0) \vert\vert < \epsilon /2$, respectively. We can take $\delta$ to be $min\{\delta_1, \delta_2\}$. Then, the following is established.
\begin{align*}
DIF &= \vert\vert (f(x)+g(x)) - (f(x_0)+g(x_0))\vert\vert\\ 
&= \vert\vert(f(x)-f(x_0)) + (g(x)-g(x_0))\vert\vert\\
&< \epsilon/2 + \epsilon/2 = \epsilon 
\end{align*}
Hence, $f+g$ is continuous.

{\bf{$f\circ g$ is continuous: }}
As $f: D_1 \rightarrow D_2$ is continuous, we have the following that for each input $u_0 \in D_1$ and  every $\epsilon > 0$ there exists an $\epsilon_0 > 0$ such that for all $u \in D_1$: $\vert\vert u-u_0 \vert\vert < \epsilon_0 \Rightarrow \vert\vert f(u)-f(u_0) \vert\vert < \epsilon$. What's more, since $g: D_0 \rightarrow D_1$ is continuous, we have the following that for each input $x_0 \in D_0$ wherein $u_0 = g(x_0)$ and  every $\epsilon_0 > 0$ there exists a $\delta > 0$ such that for all $x \in D_0$: $\vert\vert x-x_0 \vert\vert < \delta \Rightarrow \vert\vert g(x)-g(x_0) \vert\vert < \epsilon_0$, i.e., $\vert\vert u-u_0 \vert\vert < \epsilon_0$.
Therefore, we can conclude that for each input $x_0 \in D_0$ and every $\epsilon > 0$ there exists a $\delta > 0$ such that for all $x \in D_0: \vert\vert x-x_0 \vert\vert < \delta \Rightarrow \vert\vert f(g(x))-f(g(x_0)) \vert\vert < \epsilon$, which means $f\circ g$ is continuous.
\end{myproof}

For the DNNs using ReLU, each output value can be calculated from input values in terms of  ReLU function and the compositions in Theorem \ref{th2}. 
Based on the two theorems, we can directly conclude that each output value of a DNN using ReLU is continuous \emph{w.r.t.} the input values. %As we have shown, the activation function \emph{ReLU} is continuous. And in this work, the DNN we care about actually is the composition of such continuous function and some concatenation operator like $c \cdot$, $+$ and $\circ$. As the proof show above, we can speculate the DNN is continuous.
%Hence, the output space of a DNN using ReLU is continuous \emph{w.r.t.} the input space. 
According to the definitions of DNN and local robustness, an initial input with a smaller value of $N_{d}(x)$ tends to be a weak point of the DNN, because a small change of input may cause the change of $N(x)$, due to the continuity conclusion. Based on this insight, we can design a method to choose or generate the inputs with smaller $N_{d}(x)$ values; intuitively, verification tools can quickly find counter-examples around these inputs.

%!TEX root = main.tex
\section{Boosting Method}

There exist two main observations behind our boosting method: (1) there exist adversarial examples in most DNNs; and (2) the cost of finding counter-examples depends heavily on the given input. Hence, to boost the verification of DNNs, we propose a method for generating inputs that are more possible to have counter-examples around, and a greedy strategy to search the input's nearby space. Figure~\ref{framework} shows the basic framework of our boosting method.

\subsection{Framework}

%Our verification focuses on the local robustness, i.e., ensuring the safety around given inputs.  
The framework of verification contains three stages: input generation, pre-analysis, and verification. To boost the verification, we would like to evaluate an input's possibility of having counter-examples around. Hence, in the input generation stage, we use a selective sampling method to generate the  inputs that tend to violate local robustness. After that, we perform a light-weight pre-analysis that uses an efficient search heuristic to find counter-examples around the input. If we find a counter-example, we will terminate the verification and report the counter-example; otherwise, we will use the verification tool to verify the local robustness \emph{w.r.t.} the input.  
{\begin{algorithm}[!ht]
		\caption{Threshold Generation}
		\label{threshold}
		\LinesNumbered
		\DontPrintSemicolon
		$\mathsf{computeThreshold}(N, \mathcal{X})$\\
		\KwData{a DNN $N$ and a set of inputs $\mathcal{X}$}
		\Begin{
			$x\leftarrow pop(\mathcal{X})$;\\
%			$C=N(x)$;\\
			$threshold\leftarrow N_{d}(x)$;\\		
			\While{$\mathcal{X}\neq \emptyset$}{
				$x'\leftarrow pop(\mathcal{X})$; \\
%				$C'=N(x')$;\\
				$diff \leftarrow N_{d}(x')$;\\
				\If{$diff<threshold$}{ \label{accepted}
					$threshold\leftarrow diff$;\\
				}
			}
		\Return{$threshold$};
		}
\end{algorithm}}

\subsection{Input Generation}

Given a DNN $N$, we first calculate a threshold that can be used to evaluate inputs. Algorithm~\ref{threshold} shows the procedure of computing the threshold. 
We randomly generate a set of inputs, denoted as $\mathcal{X}$, Algorithm~\ref{threshold} returns  the smallest $N_d(x)$ where  $x\in\mathcal{X}$. 
%Given an input $x\in\mathcal{X}$, 
%we leverage the distance between its largest output value (denoted as $max(N(x))$) and second largest value (denoted as $\hat{max}(N(x))$) to  
%compute the threshold. 
%Algorithm~\ref{threshold} returns the smallest distance among the inputs set $\mathcal{X}$.
It is worth pointing out that we set the size of $\mathcal{X}$ to be 1000 in our experiments.
{\begin{algorithm}[!ht]
		\caption{Input Generation}
		\label{Input}
		\LinesNumbered
		\DontPrintSemicolon
		$\mathsf{getInput}(N, threshold, colNum)$\\
		\KwData{a DNN $N$, the $threshold$, and a control variable $colNum$ for random samplings}
		\Begin{		
			$num\_sampling\leftarrow0$;\\
			\While{$true$}{
				\If{$num\_sampling>colNum$}{ 
					$threshold\leftarrow threshold * 1.1$;\\
					$num\_sampling\leftarrow0$;\\
				}
			$x\leftarrow randomSampling()$;\\
%			$C=N(x)$;\\
			$diff\leftarrow N_{d}(x)$;\\
			\If{$diff<threshold$}{ 
				\Return{$x$};
			}\Else{
			  $num\_sampling {++}$;\\
	      	}
		}		
		}
\end{algorithm}}

The insight for input generation is that an input tends to have counter-examples nearby if its largest output value is close to the second largest one. 
Algorithm~\ref{Input} takes the DNN $N$, a pre-computed $threshold$, and a control variable $colNum$ as inputs, and returns an input for pre-analysis and verification. 
The idea is to use $threshold$ to evaluate a randomly generated input, \emph{i.e.}, an input $x$ is valid if $N_{d}(x)$ is smaller than $threshold$ (Lines 8-10). Note that if we fail to find a valid input under a given number of times, to reduce the difficulty in generating qualified inputs, we increase the $threshold$ gradually, \emph{i.e.}, $threshold \leftarrow threshold * 1.1$ (Lines 4-6). 

\subsection{Algorithm of Finding Counter-examples}\label{pre-analysis}

Algorithm~\ref{greedy} shows how to find a counter-example \emph{w.r.t.} a given input. The basic idea is to employ a greedy search procedure to find a counter-example. The inputs contain a DNN $N$, a given input $x$, the maximum step $L_{max}$, and the minimum step $L_{min}$. 
%We first use $x$ and  $L_{max}$ to calculate the scope (denoted as $range$) for searching counter-examples (\mbox{Line 2}). 
We set the initial $step$ to be $L_{max}/2$, and generate sample inputs by modifying each dimension of $x$ by $step$.
%within the scope, \emph{i.e.}, $\{x'\mid \vert\vert x'-x\vert\vert_{i} = step \}$. 
After that, we perform the following two checks on every sampled input. Given a sample $x'$: (1) if $x'$ is a counter-example, we report it and exit (Lines 10-11); otherwise, (2) if  $N_{d}(x')$ is smaller than that of the initial seed input $x$, we update $x$ to be $x'$ (Lines 13-17). If no counter-example is found and the seed input has not been updated, we will reduce the search scope, \emph{i.e.}, $step \leftarrow step/2$. Note that when $step$ is smaller than the minimum step, we exit directly (Lines 5-6), \emph{i.e.}, the verification tool is used to verify $x$'s local robustness. 
 
{\tiny\begin{algorithm}%[!ht]
		\caption{Algorithm of Finding Counter-examples}
		\label{greedy}
		\LinesNumbered
		\DontPrintSemicolon
		$\mathsf{greedySearch}(N,x,L_{max},L_{min})$\\
		\KwData{a DNN $N$, a given input $x$, the maximum step $L_{max}$, and the minimum step $L_{min}$}
		\Begin{
%			$range\leftarrow setRange(x,L_{max})$;\\
			$changed \leftarrow false$;\\
			$step \leftarrow L_{max}/2$;\\	
			\While{true}{
				\If{$step<L_{min}$}{ 
			         \Return{UNSAT};
				}
				$inputs\leftarrow genInputs(x,step)$; \\
				\While{$inputs\neq\emptyset$}{
				  $x'\leftarrow pop(inputs)$;\\
				  \If{$x'$ is a counter-example}{
				  	 \Return{$x'$};
				  }\Else{
				     $diff\leftarrow N_{d}(x)$;\\
				     $temp\leftarrow N_{d}(x')$;\\
				     \If{$temp<diff$}{
				     	$x\leftarrow x'$;\\
				     	$changed\leftarrow true$;\\
				     } 	
			      }
			    }
		    \If{$!changed$}{
		    	$step\leftarrow step/2$;\\
		    } 	
			}
		}
\end{algorithm}}

\subsection{Discussion}

In principle, boosting should achieve a tradeoff between boosting cost and the achieved improvement in verification. Compare with verification procedure, our boosting method is light-weight. Hence, we prefer to generate or select weak inputs to achieve more improvements in verification. Besides, if it is not feasible to generate inputs, \emph{e.g.}, image inputs, the input generation method can be used as an input selecting method.  Furthermore, it is natural that our boosting method can also be used to improve the existing methods for adversarial example generation \cite{GoodfellowSS14}, which is also validated in our evaluation.

%!TEX root = main.tex
\section{Evaluation}

%\subsection{Implementation}

In this section, we will present the evaluation of our boosting method. First, we give the research questions of evaluation. Then, the experiment setup and the results will be depicted. Last, the threats to validity will be discussed.

\subsection{Research Questions}
%To evaluate the effectiveness of our method, we designed several experiments to prove it. We have four main goals:
We have following main research questions to evaluate:
\begin{itemize}
  %\item Verify the hypothesis 2, that is, the lower the difference, the higher the probability of finding a adversarial sample around it;
%  \item Compare which method of setting the threshold is more effective;
\item \textbf{RQ1}: how effective is the boosting method for finding counter-examples during DNN verification?
%  \item Verify the selection method is effective, that is, whether the probability of SAT can be greatly improved after screening using the evaluation function;
\item \textbf{RQ2}: how effective is the pre-analysis algorithm? How dominant is it?
\item \textbf{RQ3}: how effective is the threshold generation method?
\item \textbf{RQ4}: how effective is our input selecting method for DNN attacking methods?
%  \item Prove that direct use of algorithms to find adversarial samples can improve the efficiency of Reluplex verification.
\end{itemize}

\subsection{Experiment Setup}

To answer the first three research questions, we have applied our boosting method on Reluplex \cite{KatzBDJK17}. We choose two DNN benchmarks, \emph{i.e.}, ACAS-Xu and MNIST, as our evaluation benchmarks. The DNNs in ACAS-Xu are for
 autonomous aircraft control aiming at avoiding collisions, and used by Reluplex as the benchmark. Each DNN in ACAS-Xu has 6 hidden layers, and each layer has 50 neurons. Both of input and output layers have 5 neurons. MNIST is commonly used by the existing work of DNN verification \cite{GopinathKPB18} and automatic test generation \cite{Moosavi-Dezfooli16}. Based on MNIST benchmark, we train the MNIST networks with the structure 784-100-100-10, \emph{i.e.}, the input layer has 784 neurons, the output layer has 10 neurons, and each of the two hidden layers has 100 neurons.

\begin{figure*}
  \centering
  \subfigure[Network 1]{
    \includegraphics[width=0.31\textwidth]{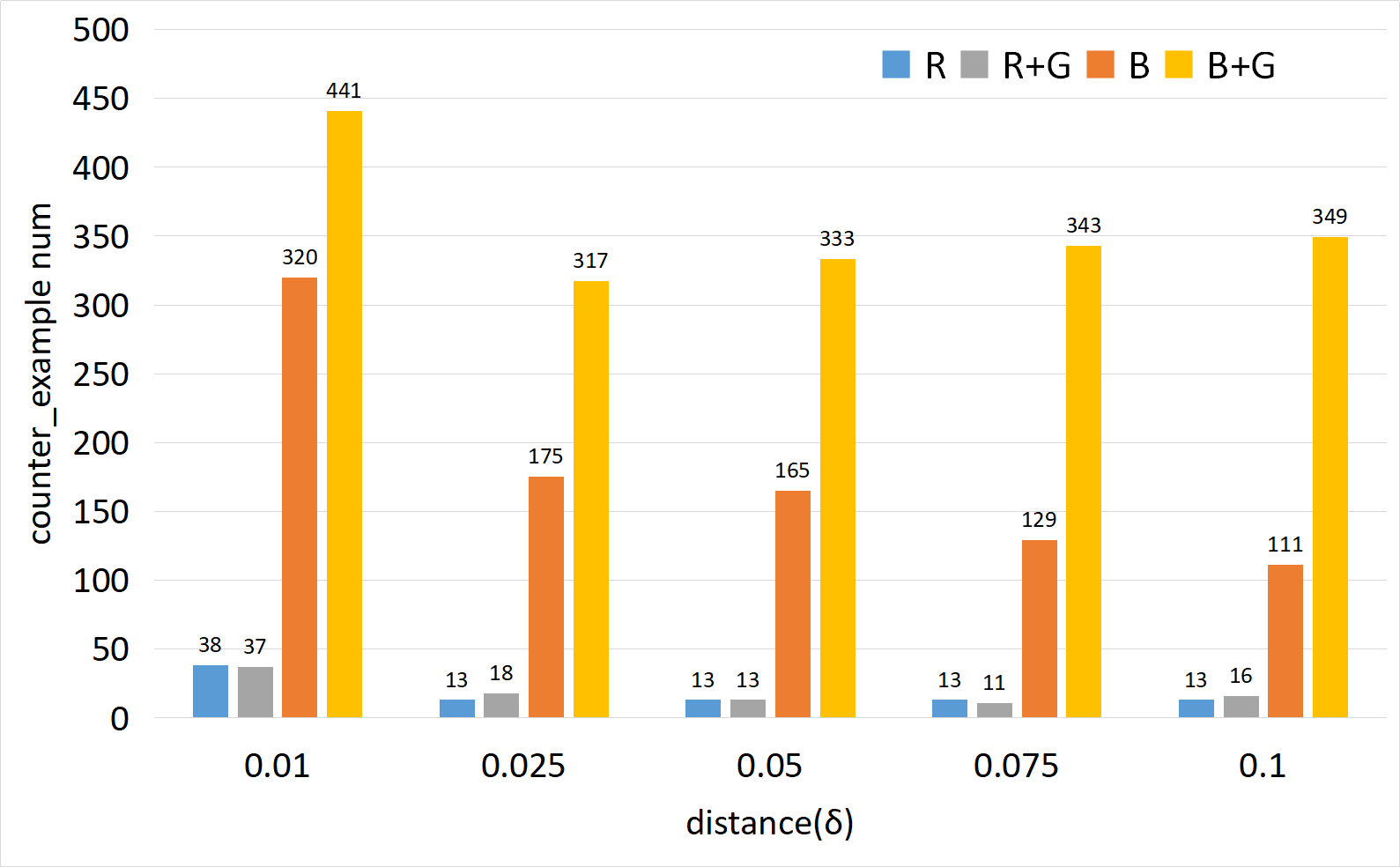}
    \label{Network 1}
  }
%  \quad
  \subfigure[Network 2]{
    \includegraphics[width=0.31\textwidth]{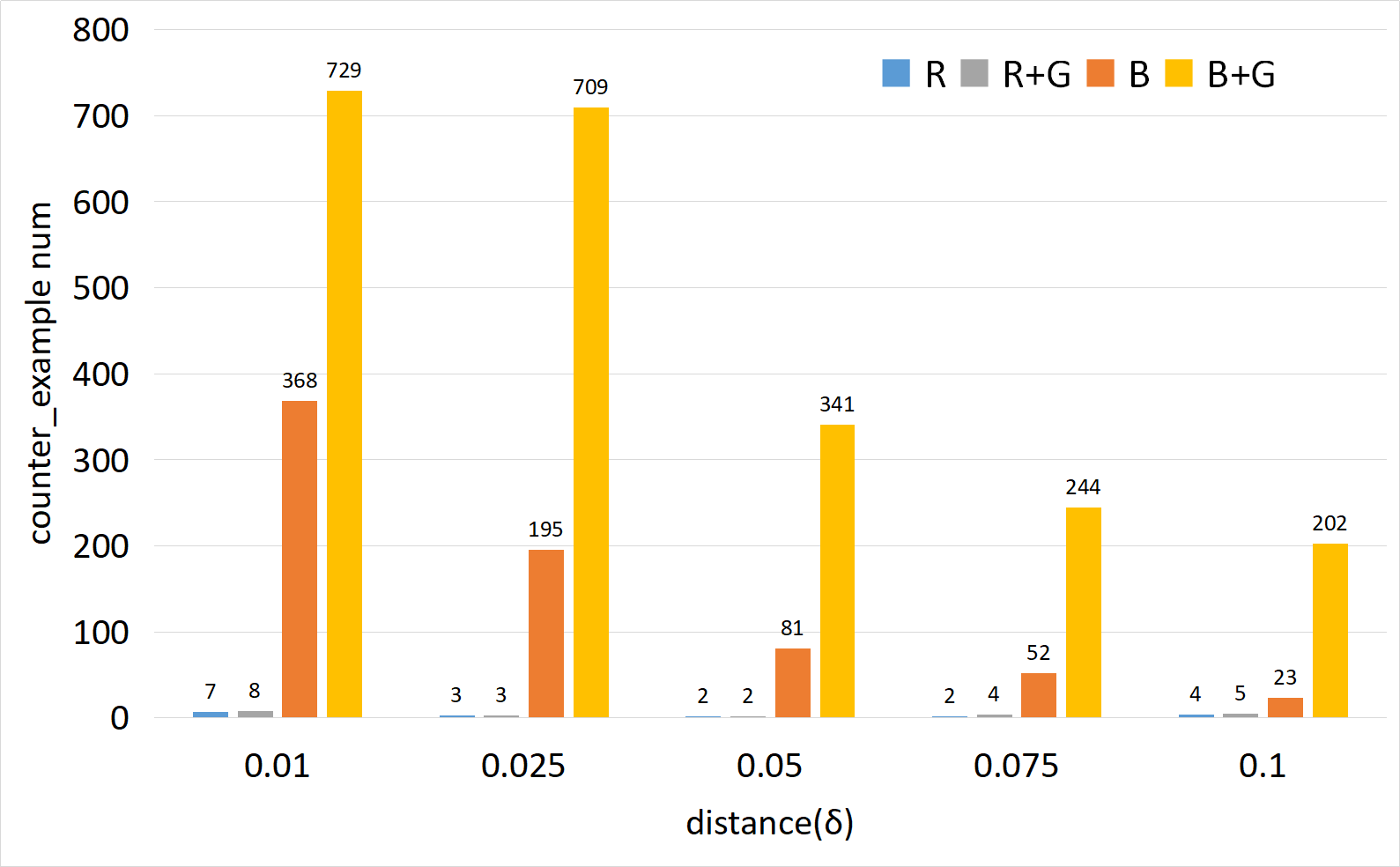}
    \label{Network 2}
  }
%  \quad
  \subfigure[Network 3]{
    \includegraphics[width=0.31\textwidth]{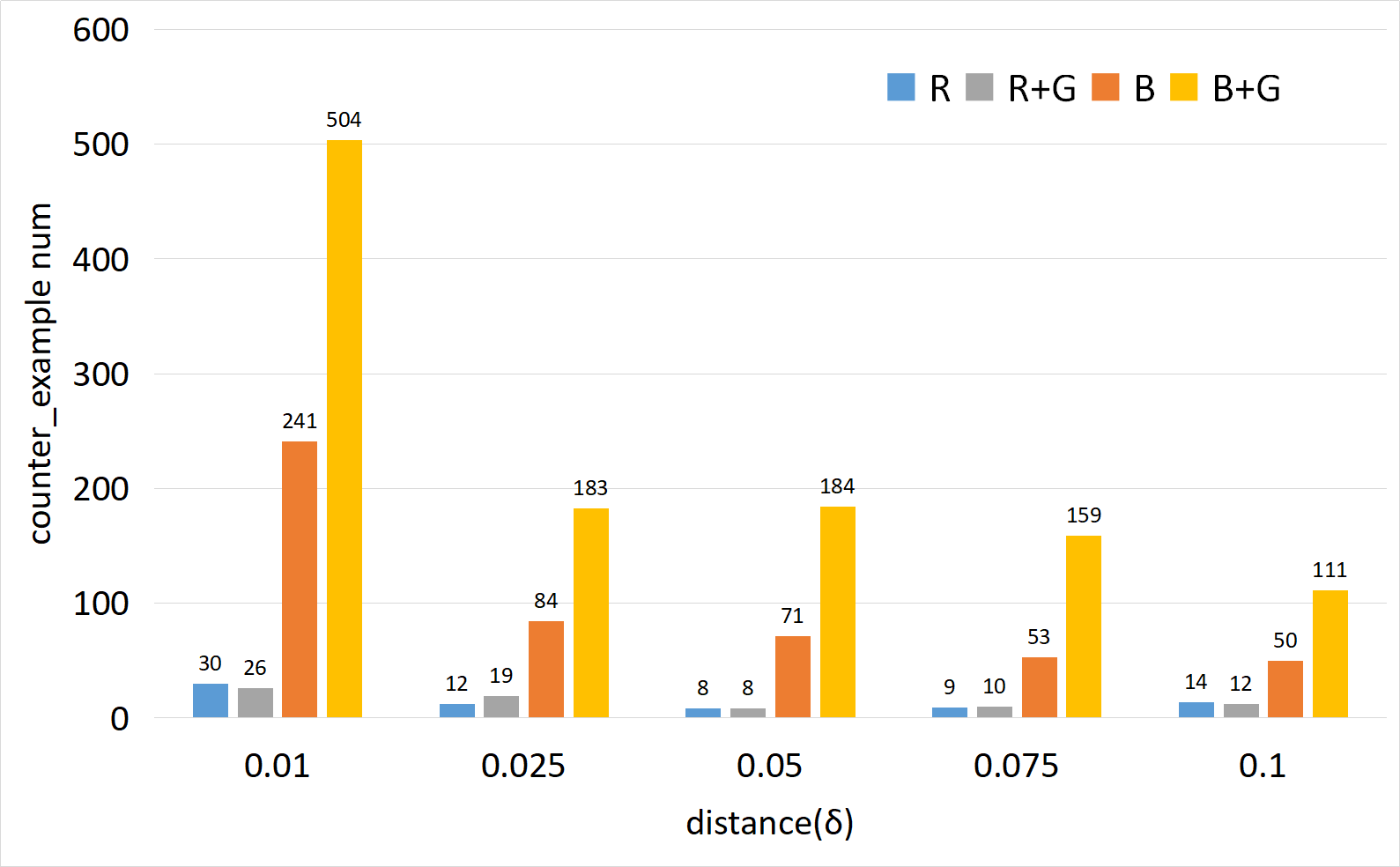}
    \label{Network 3}
  }
%  \vspace{-2mm}
  \caption{The numbers of counter-examples on ACAS-Xu networks.}
  \label{counter-examples}
\end{figure*}
\begin{figure*}
  \centering
  \subfigure[Network 1]{
    \includegraphics[width=0.31\textwidth]{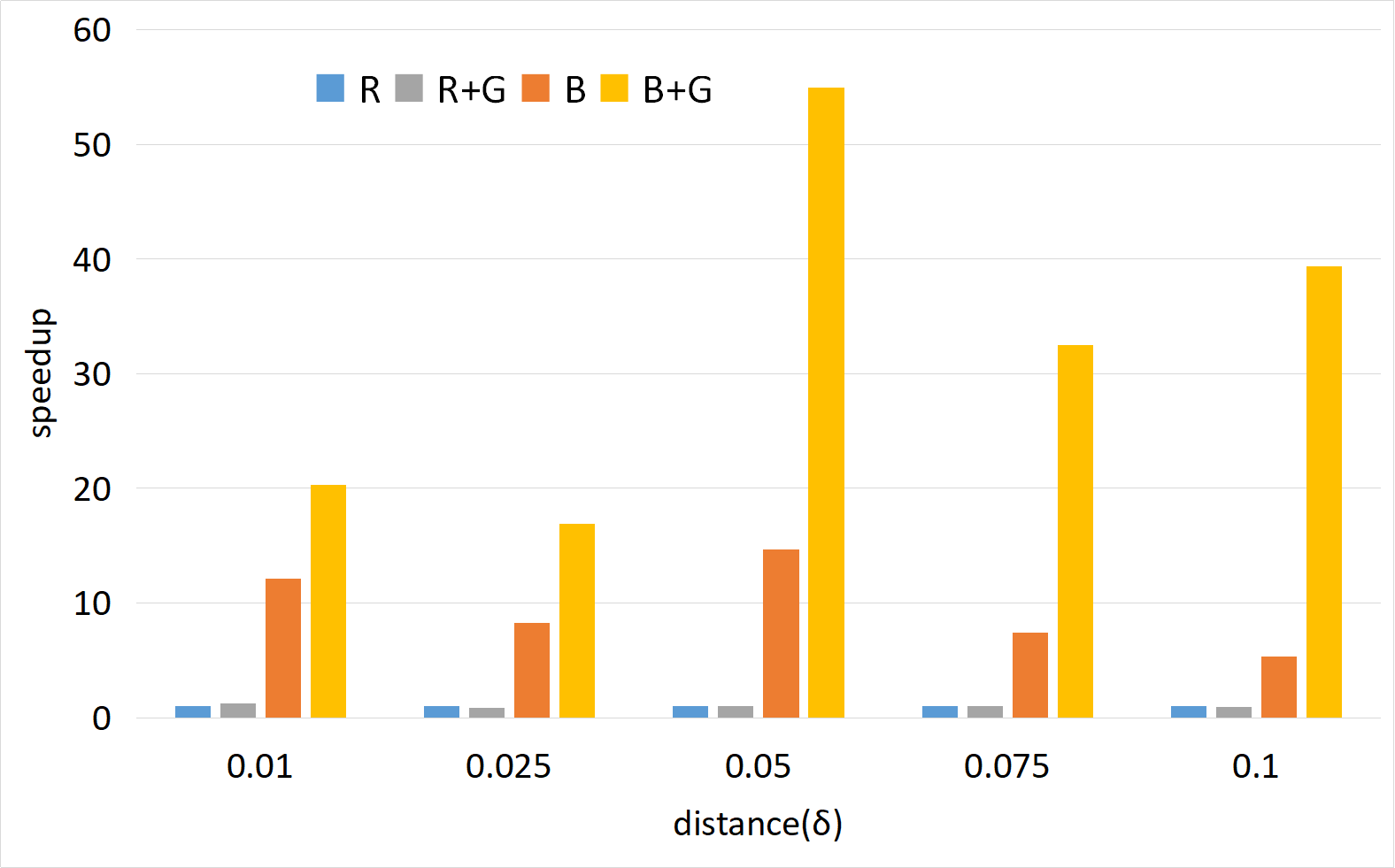}
    \label{Network 1}
  }
%  \quad
  \subfigure[Network 2]{
    \includegraphics[width=0.31\textwidth]{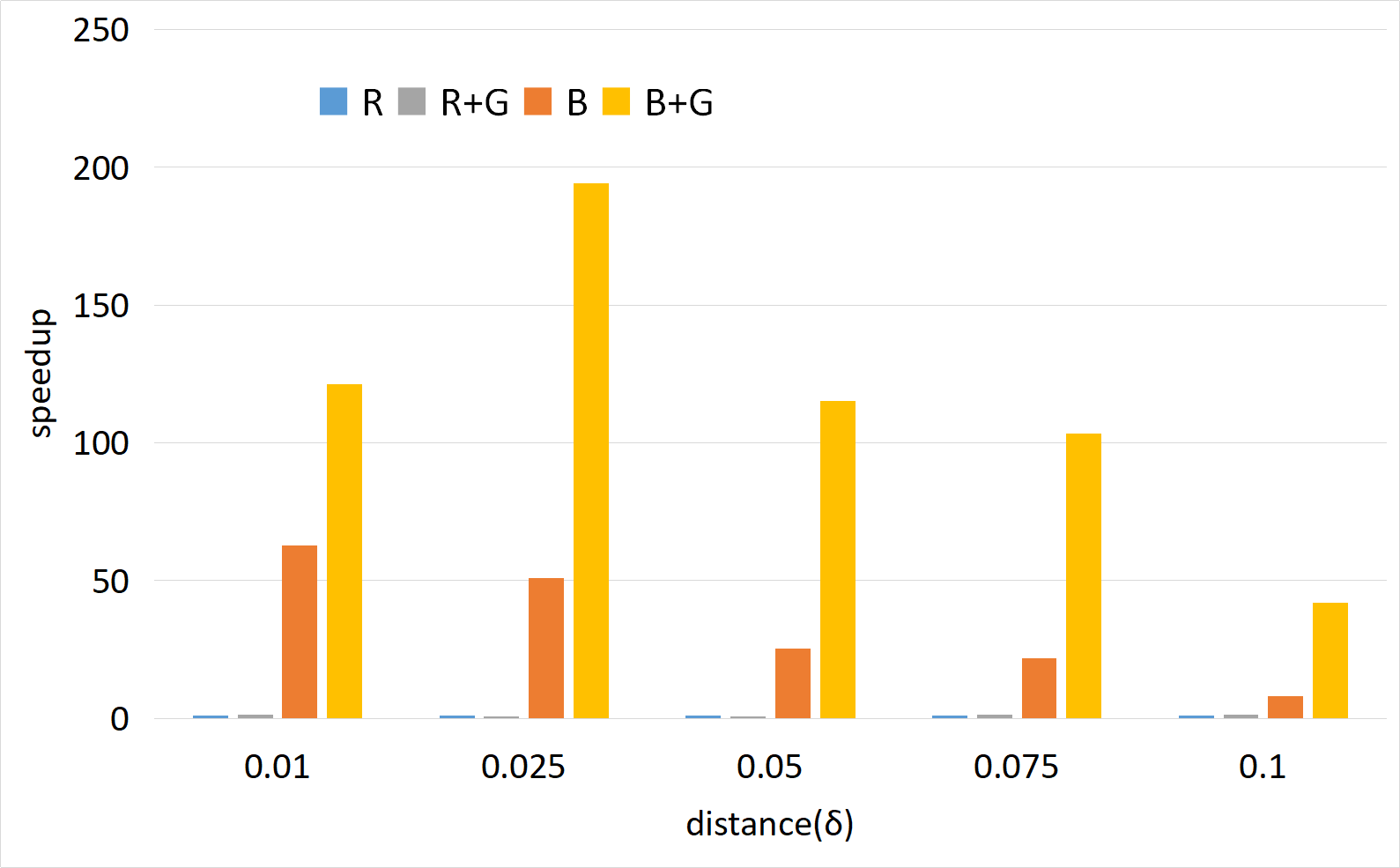}
    \label{Network 2}
  }
%  \quad
  \subfigure[Network 3]{
    \includegraphics[width=0.31\textwidth]{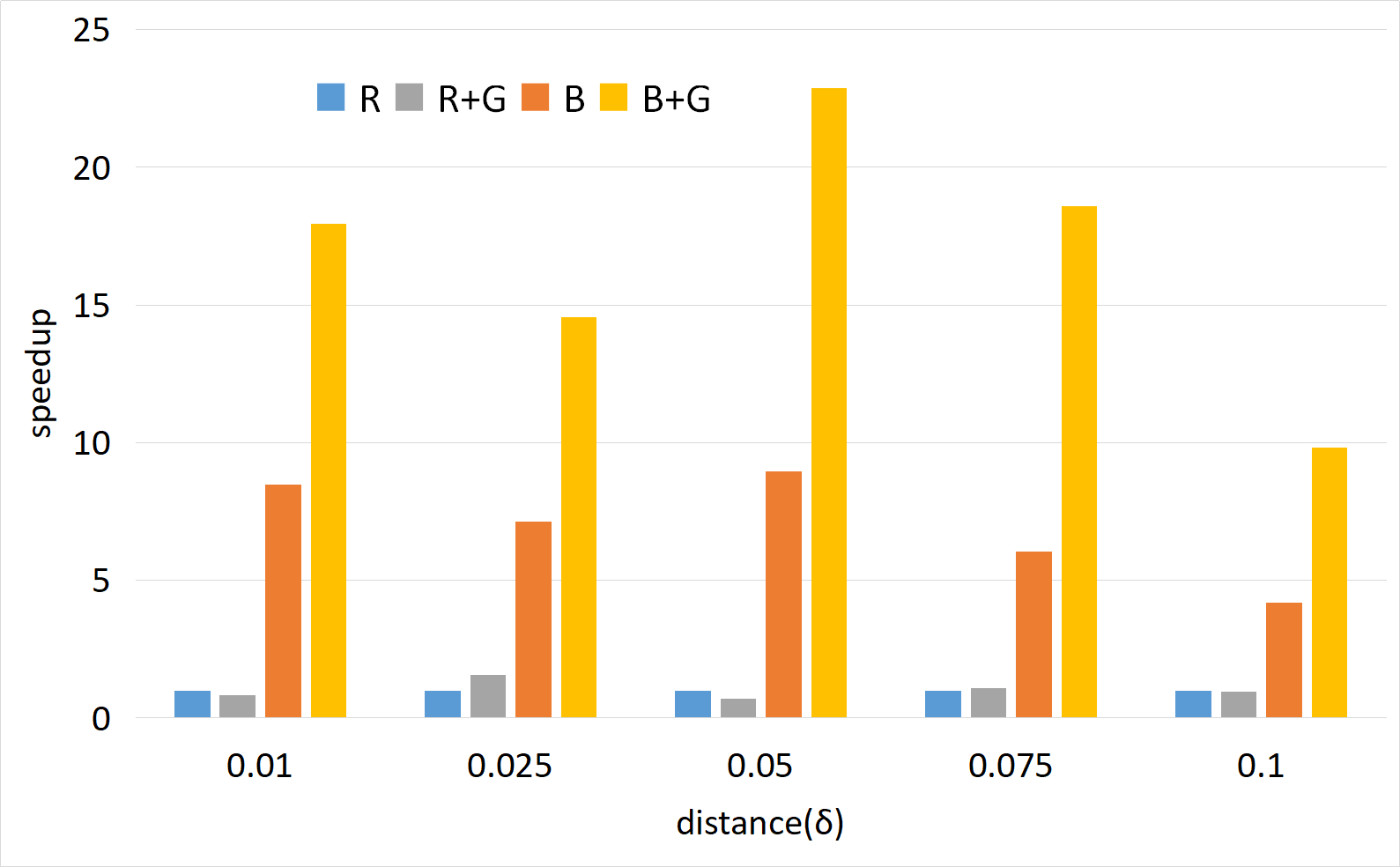}
    \label{Network 3}
  }
%  \vspace{-2mm}  
  \caption{\upshape The speedups for finding 50 counter-examples on ACAS-Xu networks.}
  \label{speedup}
\end{figure*}

%To evaluate the effectiveness of our approach in accelerating Reluplex verification, we chose the Airborne Collision Avoidance System (ACAS) dataset that comes with Reluplex. This is a neural network based advisory system recommending horizontal manoeuvres for aircraft in order to avoid collisions, based on sensor measurements\cite{XXX}. It contains five inputs, each with a continuous input range, so its test set is infinite.

We randomly select three DNNs from ACAS-Xu benchmark, and use Reluplex to verify the local robustness of the networks. Since distance $\delta$ (\emph{c.f.}, Definition \ref{local-rb-def}) greatly influences the verification result, we select five distances for evaluation. For MNIST, we trained three networks whose test accuracies are 0.9779, 0.9798 and 0.9769, respectively. We choose three distances for the evaluation on MNIST networks.

We carried out each local robustness verification task in four modes: 1) random initial input, denoted by \textsf{R}; 2) random initial input  plus greedy algorithm, denoted by \textsf{R+G}; 3) choosing initial input, denoted by \textsf{B}; 4) choosing initial point plus greedy algorithm, denoted by \textsf{B+G}. To answer the first three questions, we design two experiments on ACAS-Xu benchmark: local robustness verification in 24 hours, and finding 50 counter-examples. For MNIST networks, we only do local robustness verification in 24 hours.

To answer the last question, we combine our boosting method with four representative attacking methods: FGSM \cite{GoodfellowSS14}, JSMA \cite{PapernotMJFCS16}, CW \cite{Carlini017}, and DeepFool \cite{Moosavi-Dezfooli16}. We design an experiment on MNIST networks in which each method is used to generate adversarial examples for 1000 inputs. We compare the rate of successfully generating adversarial examples using our input selecting method and that using random input selecting. We use tensorflow-adversarial \cite{t-a} as the implementation of these attacking methods. %(Experiment setup of last Question To be added)

All the experiments were carried out on a cloud server with 64 cores (3.10GHz), 256GB memory and Ubuntu 16.04 operation system. Because the tasks were run in parallel, to eliminate experimental errors, each verification task is run three times, and we use the average results as the experimental results. Besides, because Reluplex often needs a long time for verification, we set the time limit of single local robustness verification as 20 minutes for ACAS-Xu networks and 60 minutes for MNIST networks. If the time limit is reached and no result is obtained, we stop the single run.

%To illustrate the generality, we selected five neural networks for verification, and each experiment was run three times to reduce the mistake. Our method does not depend on the scale of the DNN, but the calculation of the large DNN on the Reluplex is impossible, so it is difficult to verify the improvement of our method on large DNN.

%Since different $\delta$ have a great influence on the Reluplex verification time, we selected five $\delta$ to verify them separately. At the same time, because the verification time of individual singular points will greatly affect the whole experiment (for example, the verification time of one point is more than the verification time of all other points), we set the time $T$ to cut off. When the running time exceeds $T$, the point is considered to be an UNSAT, and the verification of this point is stopped.
\begin{table}[!ht]
  \centering
  \caption{ Experimental Results on MNIST networks.}
  {\small\begin{tabular}{c|c|c|ccc|c}
  \hline
  \multirow{2}{*}{Network} &   \multirow{2}{*}{$\delta$} &   \multirow{2}{*}{Mode} & \multicolumn{4}{c}{24 hours robustness verification}\\
  \cline{4-7}
  &&& \#Runs & \#SAT$^1$ & \#SAT$^2$ & Rate \\
  \hline
  \multirow{12}{*}{1} & \multirow{ 4}{*}{2} & \textsf{R} &  6 & 0 & 0 & 0.00\%\\
   &  & \textsf{R+G} &  6 & 0 & 0 & 0.00\% \\
   &  & \textsf{B} &  94 & 91 & 0 & 96.81\%\\
   &  & \textsf{B+G} &  207 & 201 & 120 & 97.10\% \\
  \cline{2-7}
   & \multirow{ 4}{*}{6} & \textsf{R} &  3 & 0 & 0 & 0.00\% \\
   &  & \textsf{R+G} &  3 & 0 & 0 & 0.00\% \\
   &  & \textsf{B} &  45 & 42 & 0 & 93.33\% \\
   &  & \textsf{B+G} &  104 & 97 & 59 & 93.27\% \\
  \cline{2-7}
   & \multirow{ 4}{*}{12} & \textsf{R} &  3 & 0 & 0 & 0.00\%\\
   &  & \textsf{R+G} &  3 & 0 & 0 & 0.00\%\\
   &  & \textsf{B} &  32 & 23 & 0 & 71.88\%\\
   &  & \textsf{B+G} &  80 & 67 & 48 & 83.75\%\\
  \hline
  \multirow{12}{*}{2} & \multirow{ 4}{*}{2} & \textsf{R} & 9 & 0 & 0 & 0.00\%\\
   &  & \textsf{R+G} &  9 & 0 & 0 & 0.00\%\\
   &  & \textsf{B} &  93 & 89 & 0 & 95.70\%\\
   &  & \textsf{B+G} &  169 & 165 & 79 & 97.63\%\\
  \cline{2-7}
   & \multirow{ 4}{*}{6} & \textsf{R} &  3 & 0 & 0 & 0.00\%\\
   &  & \textsf{R+G} &  3 & 1 & 1 & 33.33\%\\
   &  & \textsf{B} &  46 & 38 & 0 & 82.61\%\\
   &  & \textsf{B+G} &  80 & 72 & 38 & 90.00\%\\
  \cline{2-7}
   & \multirow{ 4}{*}{12} & \textsf{R} &  2 & 0 & 0 & 0.00\%\\
   &  & \textsf{R+G} &  2 & 0 & 0 & 0.00\%\\
   &  & \textsf{B} &  41 & 35 & 0 & 85.37\%\\
   &  & \textsf{B+G} &  64 & 54 & 27 & 84.38\%\\
  \hline
  \multirow{12}{*}{3} & \multirow{ 4}{*}{2} & \textsf{R} &  8 & 1 & 0 & 12.50\%\\
   &  & \textsf{R+G} &  9 & 1 & 1 & 11.11\%\\
   &  & \textsf{B} &  85 & 84 & 0 & 98.82\%\\
   &  & \textsf{B+G} &  145 & 144 & 69 & 99.31\%\\
  \cline{2-7}
   & \multirow{ 4}{*}{6} & \textsf{R} &  3 & 0 & 0 & 0.00\%\\
   &  & \textsf{R+G} &  4 & 1 & 1 & 25.00\%\\
   &  & \textsf{B} &  44 & 41 & 0 & 93.18\%\\
   &  & \textsf{B+G} &  86 & 82 & 42 & 95.35\%\\
  \cline{2-7}
   & \multirow{ 4}{*}{12} & \textsf{R} &  3 & 0 & 0 & 0.00\%\\
   &  & \textsf{R+G} &  3 & 0 & 0 & 0.00\%\\
   &  & \textsf{B} &  31 & 20 & 0 & 64.52\%\\
   &  & \textsf{B+G} &  64 & 53 & 32 & 82.81\%\\
  \hline
  \end{tabular}
  \label{result_mnist}
  }
\end{table}

%!TEX root = main.tex
\begin{table*}
  \centering
  \caption{ Experimental Results on ACAS-Xu networks.}
  {\small\begin{tabular}{c|c|c|rrr|r|rrr|r}
  \hline
  \multirow{2}{*}{Network} &   \multirow{2}{*}{$\delta$} &   \multirow{2}{*}{Mode} & \multicolumn{4}{c|}{24 hours robustness verification} & \multicolumn{4}{c}{Finding 50 counter-examples}\\
  \cline{4-11}
  &&& \#Runs & \#SAT$^1$ & \#SAT$^2$ & Rate & \#Runs & \#SAT$^2$ & Time(mins) & Rate\\
  \hline
  \multirow{20}{*}{1} & \multirow{ 4}{*}{0.01} & \textsf{R} &  244 & 38 & 0 & 15.71\% & 348 & 0 & 2374.67 & 14.35\%\\
   &  & \textsf{R+G} &  259 & 37 & 21 & 14.14\% & 324 & 22 & 1934 & 15.42\%\\
   &  & \textsf{B} &  361 & 320 & 0 & 88.56\% & 58 & 0 & 196.33 & 86.21\%\\
   &  & \textsf{B+G} &  512 & 441 & 198 & 86.20\% & 55 & 22 & 116.67 & 91.46\%\\
  \cline{2-11}
   & \multirow{ 4}{*}{0.025} & \textsf{R} &  81 & 13 & 0 & 16.53\% & 221 & 0 & 3914 & 22.62\%\\
   &  & \textsf{R+G} &  77 & 18 & 9 & 22.94\% & 250 & 29 & 4364.67 & 20.03\%\\
   &  & \textsf{B} &  184 & 175 & 0 & 95.46\% & 52 & 0 & 473 & 96.77\%\\
   &  & \textsf{B+G} &  326 & 317 & 193 & 97.14\% & 51 & 32 & 230.67 & 97.40\%\\
  \cline{2-11}
   & \multirow{ 4}{*}{0.05} & \textsf{R} &  35 & 13 & 0 & 36.54\% & 167 & 0 & 6810.33 & 29.94\%\\
   &  & \textsf{R+G} &  41 & 13 & 7 & 31.71\% & 167 & 27 & 6577.33 & 29.94\%\\
   &  & \textsf{B} &  168 & 165 & 0 & 98.02\% & 52 & 0 & 463 & 95.54\%\\
   &  & \textsf{B+G} &  342 & 333 & 234 & 97.37\% & 50 & 35 & 124.33 & 100\%\\
  \cline{2-11}
   & \multirow{ 4}{*}{0.075} & \textsf{R} &  34 & 13 & 0 & 39.60\% & 134 & 0 & 6370.67 & 37.31\%\\
   &  & \textsf{R+G} &  31 & 11 & 6 & 36.96\% & 155 & 34 & 855.67 & 32.33\%\\
   &  & \textsf{B} &  133 & 129 & 0 & 96.98\% & 54 & 0 & 855.67 & 92.59\%\\
   &  & \textsf{B+G} &  348 & 343 & 228 & 98.66\% & 51 & 33 & 196.33 & 98.68\%\\
  \cline{2-11}
   & \multirow{ 4}{*}{0.1} & \textsf{R} &  30 & 13 & 0 & 42.70\% & 107 & 0 & 5078.67 & 46.73\%\\
   &  & \textsf{R+G} &  34 & 16 & 11 & 45.63\% & 124 & 26 & 5510.33 & 40.43\%\\
   &  & \textsf{B} &  117 & 111 & 0 & 95.14\% & 55 & 0 & 952.67 & 90.91\%\\
   &  & \textsf{B+G} &  356 & 349 & 251 & 97.94\% & 50 & 35 & 129 & 100\%\\
  \hline
  \multirow{20}{*}{2} & \multirow{ 4}{*}{0.01} & \textsf{R} & 177 & 7 & 0 & 3.77\% &  1472 & 0 & 12236 & 3.40\%\\
   &  & \textsf{R+G} &  189 & 8 & 5 & 4.23\% &  1186 & 32 & 9124.33 & 4.21\%\\
   &  & \textsf{B} &  387 & 368 & 0 & 95.26\% &  52 & 0 & 195.33 & 96.15\%\\
   &  & \textsf{B+G} &  751 & 729 & 536 & 97.07\% &  51 & 36 & 101 & 98.04\%\\
  \cline{2-11}
   & \multirow{ 4}{*}{0.025} & \textsf{R} &  37 & 3 & 0 & 19.05\% &  517 & 0 & 19804 & 9.68\%\\
   &  & \textsf{R+G} &  43 & 3 & 1 & 7.03\% &  711 & 15 & 24127.67 & 7.03\%\\
   &  & \textsf{B} &  200 & 195 & 0 & 97.67\% &  51 & 0 & 388.33 & 97.40\%\\
   &  & \textsf{B+G} &  711 & 710 & 546 & 99.81\% &  50 & 38 & 101.67 & 100\%\\
  \cline{2-11}
   & \multirow{ 4}{*}{0.05} & \textsf{R} &  24 & 2 & 0 & 9.72\% &  508 & 0 & 27158.33 & 9.84\%\\
   &  & \textsf{R+G} &  24 & 2 & 1 & 6.94\% &  600 & 23 & 37091.67 & 8.33\%\\
   &  & \textsf{B} &  93 & 81 & 0 & 87.41\% &  59 & 0 & 1068 & 84.75\%\\
   &  & \textsf{B+G} &  353 & 341 & 280 & 96.60\% &  52 & 41 & 236.33 & 96.77\%\\
  \cline{2-11}
   & \multirow{ 4}{*}{0.075} & \textsf{R} &  20 & 2 & 0 & 10.00\% &  475 & 0 & 31644.33 & 10.53\%\\
   &  & \textsf{R+G} &  22 & 4 & 1 & 16.67\% &  338 & 19 & 22922.33 & 14.78\%\\
   &  & \textsf{B} &  66 & 52 & 0 & 78.89\% &  63 & 0 & 1452.33 & 78.95\%\\
   &  & \textsf{B+G} &  258 & 244 & 202 & 94.45\% &  53 & 41 & 306 & 94.94\%\\
  \cline{2-11}
   & \multirow{ 4}{*}{0.1} & \textsf{R} &  22 & 4 & 0 & 16.92\% &  350 & 0 & 24253.33 & 14.29\%\\
   &  & \textsf{R+G} &  23 & 5 & 2 & 21.74\% &  287 & 24 & 19539.33 & 17.42\%\\
   &  & \textsf{B} &  39 & 23 & 0 & 59.48\% &  81 & 0 & 2988 & 61.98\%\\
   &  & \textsf{B+G} &  216 & 202 & 169 & 93.36\% &  56 & 40 & 577 & 89.29\%\\
  \hline
  \multirow{20}{*}{3} & \multirow{ 4}{*}{0.01} & \textsf{R} &  175 & 30 & 0 & 16.92\% &  296 & 0 & 2564 & 16.91\%\\
   &  & \textsf{R+G} &  222 & 26 & 9 & 11.58\% &  446 & 17 & 3053 & 11.22\%\\
   &  & \textsf{B} &  249 & 241 & 0 & 96.92\% &  52 & 0 & 303 & 96.77\%\\
   &  & \textsf{B+G} &  509 & 504 & 251 & 99.08\% &  50 & 24 & 143 & 99.34\%\\
  \cline{2-11}
   & \multirow{ 4}{*}{0.025} & \textsf{R} &  63 & 12 & 0 & 19.05\% &  268 & 0 & 6158.33 & 18.63\%\\
   &  & \textsf{R+G} &  73 & 19 & 9 & 26.48\% &  194 & 24 & 3938.67 & 25.73\%\\
   &  & \textsf{B} &  85 & 84 & 0 & 99.61\% &  50 & 0 & 861.67 & 100\%\\
   &  & \textsf{B+G} &  187 & 183 & 95 & 97.51\% &  52 & 25 & 422.67 & 96.77\%\\
  \cline{2-11}
   & \multirow{ 4}{*}{0.05} & \textsf{R} &  29 & 8 & 0 & 27.27\% &  185 & 0 & 9280.33 & 27.08\%\\
   &  & \textsf{R+G} &  31 & 8 & 3 & 25.81\% &  264 & 19 & 13169.67 & 18.94\%\\
   &  & \textsf{B} &  73 & 71 & 0 & 96.36\% &  51 & 0 & 1038.33 & 97.40\%\\
   &  & \textsf{B+G} &  185 & 184 & 113 & 99.10\% &  50 & 30 & 406.33 & 99.34\%\\
  \cline{2-11}
   & \multirow{ 4}{*}{0.075} & \textsf{R} &  27 & 9 & 0 & 32.50\% &  157 & 0 & 8646.33 & 31.78\%\\
   &  & \textsf{R+G} &  28 & 10 & 5 & 35.71\% &  147 & 24 & 7931.33 & 33.94\%\\
   &  & \textsf{B} &  57 & 53 & 0 & 92.40\% &  54 & 0 & 1428.67 & 93.17\%\\
   &  & \textsf{B+G} &  163 & 159 & 94 & 97.75\% &  51 & 29 & 464.67 & 98.68\%\\
  \cline{2-11}
   & \multirow{ 4}{*}{0.1} & \textsf{R} &  29 & 14 & 0 & 47.13\% &  115 & 0 & 6037.67 & 43.60\%\\
   &  & \textsf{R+G} &  28 & 12 & 6 & 42.35\% &  122 & 24 & 6404.33 & 40.87\%\\
   &  & \textsf{B} &  55 & 50 & 0 & 90.85\% &  55 & 0 & 1445.67 & 90.91\%\\
   &  & \textsf{B+G} &  117 & 111 & 65 & 94.32\% &  53 & 31 & 615 & 94.34\%\\
  \hline
  \end{tabular}
  \label{result1}
  }
\end{table*}

\subsection{Experimental Results}\label{exp}

Table \ref{result1} and Table \ref{result_mnist} show the experimental results on ACAS-Xu and MNIST networks, respectively. In Table \ref{result1}, the first column is the network index. The second column $\delta$ is the distance value. The third column shows the experiment mode.
Then, the rest two big columns show the results of the two experiments, \emph{i.e.}, 24 hours verification and finding 50 counter-examples, respectively. In the first big column, column \#Runs shows the total runs in 24 hours; column \#SAT$^1$ shows the total number of found counter-examples; column \#SAT$^2$ shows the total number of counter-examples found by greedy algorithm; and column Rate shows the rate of finding counter-examples, \emph{i.e.}, \#SAT$^1$/\#Runs. In the second big column, column \#Runs shows the total number of runs for finding 50 counter-examples;  column \#SAT$^2$ shows the total number of counter-examples found by greedy algorithm; column Time(mins) shows the time needed for finding 50 counter-examples in minutes. In Table \ref{result_mnist}, each column has the same meaning of the column with the same name in Table \ref{result1}.

\subsubsection{Boosting method's effectiveness for DNN verification}

As indicated by Tables \ref{result1}\&\ref{result_mnist}, for each case, \textsf{B+G} finds more counter-examples in 24 hours. Meanwhile, for finding 50 counter-examples, \textsf{B+G} needs the minimum time for verifying ACAS-Xu networks. These global results indicate the effectiveness of our boosting method.

%\begin{figure}[!ht]
%  \centering
%  \includegraphics[width=0.4\textwidth]{SAT_probability_0_01.png}
%  \caption{Total number of counter-examples on ACAS-Xu benchmark}
%  \label{counter-examples}
%\end{figure}

For ACAS-Xu benchmark, Figure \ref{counter-examples} shows the total numbers of counter-examples in 24 hours of different networks under different distances. As shown by the figure, \textsf{B+G} finds the largest amount of counter-examples on all distances. In average, compared with \textsf{R}, \textsf{R+G} and \textsf{B}, \textsf{B+G} can find 22.3, 19.8 and 1.9 times more counter-examples in 24 hours  on ACAS-Xu networks. Figure \ref{speedup} shows the time speedups for finding 50 counter-examples on each ACAS-Xu network under different distances, and the baseline is the time under \textsf{R} mode. As shown by the figure, \textsf{R} and \textsf{R+G} have a competitive efficiency. Both of \textsf{B} and \textsf{B+G} can significantly boost finding 50 counter-examples.
The speedups of \textsf{B} range from 4.2 to 62.7, and the speedups of \textsf{B+G} range from 9.8 to 194.2. Besides, the speedups are different on different networks. Compared with \textsf{R}, \textsf{R+G} and \textsf{B}, \textsf{B+G} has the average  23.1, 20.6 and 2.1 time speedups, respectively. %Hence, these results also demonstrate the effectiveness of our boosting method.

\begin{figure}[!ht]
  \centering
  \includegraphics[width=0.45\textwidth]{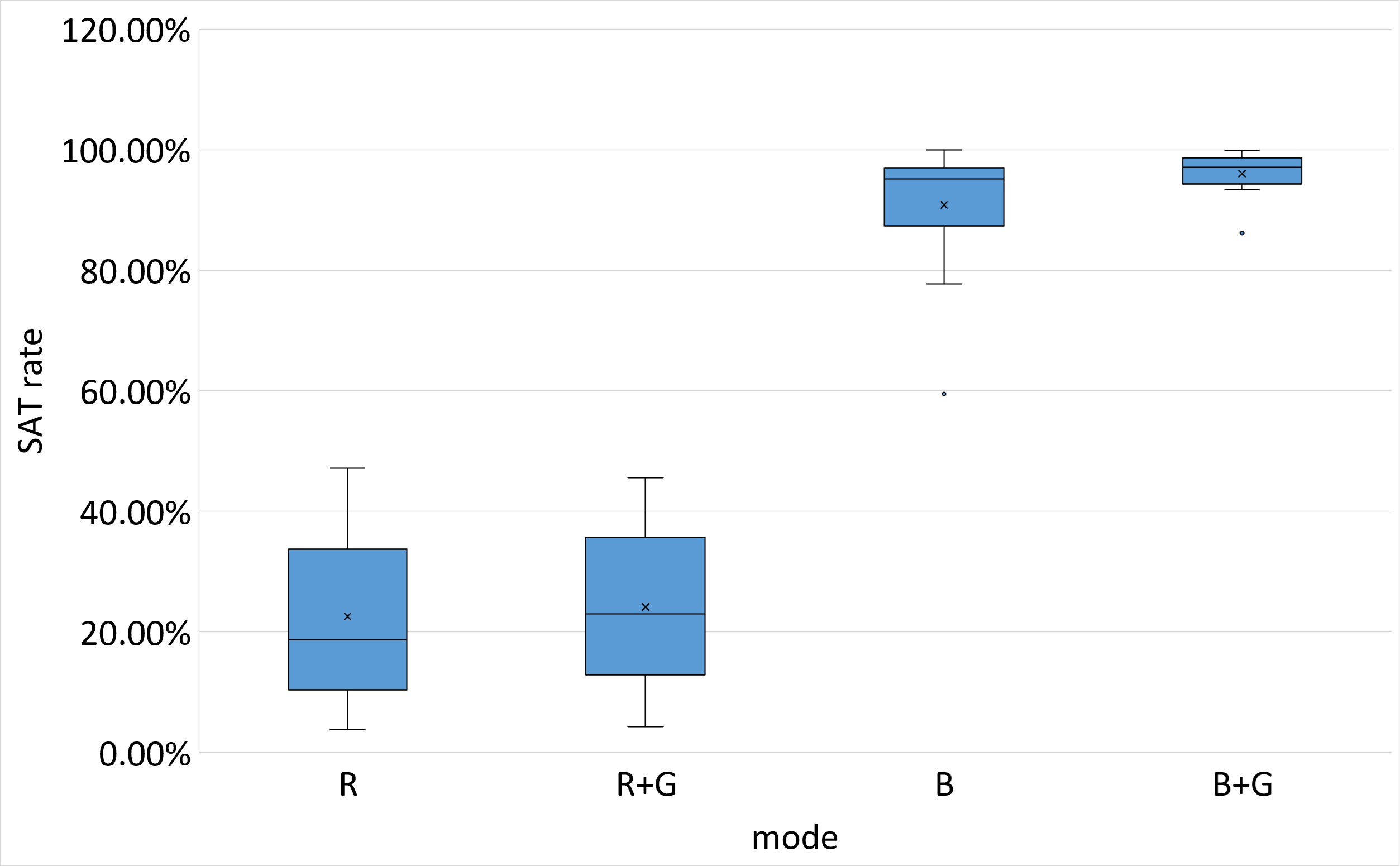}
%    \vspace{-2mm}
  \caption{\upshape Rate of finding counter-examples on ACAS-Xu networks.}
  \label{rate}
\end{figure}

Figure \ref{rate} shows the rates of finding counter-examples under different distances on ACAS-Xu benchmark. As shown by the figure, the rates of \textsf{B+G} and \textsf{B} are higher than those of \textsf{R} and \textsf{R+G}. In average, the rates of \textsf{B+G}, \textsf{B}, \textsf{R+G} and \textsf{R} are 96.47\%, 91.90\%, 25.14\%, 23.73\%, respectively. It indicates that our boosting method significantly improves the rate of finding counter-examples.

As shown in Table \ref{result_mnist}, \textsf{B+G} performs even better on MNIST benchmark. On MNIST networks, using random method (\textsf{R}) fails to find any counter-examples in most cases, even with greedy algorithms (\textsf{R+G}). The reason is MNIST networks have a large number of input variables, \emph{i.e.}, 784, which makes symbolic solving time-consuming. However, \textsf{B} and \textsf{B+G} have the average rates of 86.91\% and 91.51\%, respectively. It indicates that our boosting method greatly improves the efficiency of finding counter-examples in MNIST network verification.

%\begin{table}
%  \centering
%  \caption{\upshape The time reduce while find the same num of SAT input , and the SAT num increase while run the same time}
%  \begin{tabular}{cc|cc}
%  \hline
%  % after \\: \hline or \cline{col1-col2} \cline{col3-col4} ...
%  network & $\delta$ & Time reduce & SAT num increase \\
%  \hline
%  1 & 0.01 & 4.91\% & 1051.30\% \\
%  1 & 0.025 & 5.89\% & 2275.00\% \\
%  1 & 0.05 & 1.83\% & 2526.32\% \\
%  1 & 0.075 & 3.08\% & 2472.50\% \\
%  1 & 0.1 & 2.54\% & 2652.63\% \\
%  \hline
%  2 & 0.01 & 0.83\% & 10835.00\% \\
%  2 & 0.025 & 0.51\% & 23544.44\% \\
%  2 & 0.05 & 0.87\% & 14500.00\% \\
%  2 & 0.075 & 0.97\% & 12100.00\% \\
%  2 & 0.1 & 2.38\% & 5400.00\% \\
%  \hline
%  3 & 0.01 & 5.58\% & 1598.88\% \\
%  3 & 0.025 & 6.86\% & 1422.22\% \\
%  3 & 0.05 & 4.38\% & 2195.83\% \\
%  3 & 0.075 & 5.37\% & 1734.62\% \\
%  3 & 0.1 & 10.19\% & 709.76\% \\
%  \hline
%  4 & 0.01 & 2.19\% & 4661.90\% \\
%  4 & 0.025 & 1.75\% & 7237.50\% \\
%  4 & 0.05 & 4.61\% & 1927.78\% \\
%  4 & 0.075 & 5.93\% & 1442.31\% \\
%  4 & 0.1 & 9.65\% & 850.00\% \\
%  \hline
%  5 & 0.01 & 3.84\% & 3371.88\% \\
%  5 & 0.025 & 2.13\% & 4835.71\% \\
%  5 & 0.05 & 1.57\% & 4675.00\% \\
%  5 & 0.075 & 5.07\% & 1444.44\% \\
%  5 & 0.1 & 2.53\% & 1770.00\% \\
%  \hline
%  \end{tabular}
%
%  \label{result}
%\end{table}
%\vspace{-2mm}
\begin{framed}
\noindent \emph{Answer to RQ1:
Our boosting method can significantly improve the efficiency of finding counter-examples in DNN local robustness verification. In average, the rate of finding counter-examples using our method is more than 90\% on ACAS-Xu and MNIST networks.
}\end{framed}

%\paragraph{\bf{SAT probability}}\textcolor{red}{Figure \ref{SAT}} shows the increase of SAT probability that using the single improved method or the both improved method compared to random method. The experimental results show that the direct use of the algorithm to find the adversarial samples is not helpful to increase the SAT probability. While selecting an input with a smaller evaluation value has a greatly improving of the SAT probability. When using two improved methods at the same time, the SAT probability can be increased to nearly 100\% in each case. Therefore, in comparison, when the neural network has a lower SAT probability under the random method, our method has better improvement effect on it.

%\paragraph{\bf{Effectiveness}}To illustrate the improvement of verification efficiency in our method, we designed the experiment from two aspects: comparing the run time of the two methods(both two improved method, random method) that found the same number of SAT inputs; comparing the number of SAT inputs found by the two methods in the same time. For the first experiment, we set the number of SAT inputs that need to be found to be 50. For the second experiment, we set the run time to 24 hours. The results are shown in \textcolor{red}{Table \ref{result}}. Our method has different degrees of improvement for different neural networks, but basically can achieve a 20-fold improvement, which shows that our method is effective.

\subsubsection{Effectiveness and dominance of pre-analysis method}

We also inspect the effectiveness and dominance of pre-analysis method, \emph{i.e.}, the greedy algorithm. Table \ref{greedyResult} shows the experimental results on ACAS-Xu networks. Column (\#SAT$^1_{\textsf{X+G}}$-\#SAT$^1_{\textsf{X}}$)/\#SAT$^1_{\textsf{X}}$ shows the effectiveness of the greedy algorithm on modes \textsf{R} and \textsf{B}. Column \#SAT$^2$/\#SAT$^1$ shows the dominance of the greedy algorithm.

%\begin{figure}[!ht]
%  \centering
%  \includegraphics[width=0.4\textwidth]{SAT_probability_0_01.png}
%  \caption{Counter-examples by pre-analysis}
%  \label{rate}
%\end{figure}
%\vspace{-2mm}
\begin{table}[!ht]
  \centering
  \caption{The effectiveness and dominance of Greedy Algorithm On ACAX-Xu networks.}
%  \vspace{-2mm}
  {\begin{tabular}{c|c|cc|cc}
  \hline
  \multirow{2}{*}{Network} &   \multirow{2}{*}{$\delta$} & \multicolumn{2}{c|}{(\#SAT$^1_{\textsf{X+G}}$\ -\ \#SAT$^1_{\textsf{X}}$)/\#SAT$^1_{\textsf{X}}$} & \multicolumn{2}{c}{\#SAT$^2$/\#SAT$^1$}\\
  \cline{3-6}
  && \textsf{R} & \textsf{B} & \textsf{R+G} & \textsf{B+G}\\
  \hline
  \multirow{ 5}{*}{1} & 0.01  & -4.35\% & 37.92\% &  56.36\% & 44.79\%\\
   & 0.025 & 32.50\% & 80.61\% &  49.06\% & 60.84\%\\
   & 0.05 & 2.63\% & 101.62\% &  56.41\% & 70.24\%\\
   & 0.075 & -15.00\% & 166.58\% &  52.94\% & 66.38\%\\
   & 0.1 & 23.66\% & 214.11\% &  70.21\% & 71.99\%\\
  \hline
  \multirow{ 5}{*}{2} & 0.01 & 20.00\% & 97.92\% &  66.67\% & 73.53\%\\
   & 0.025 & 0.00\% & 263.14\% &  22.22\% & 76.97\%\\
   & 0.05 & -28.57\% & 320.58\% &  40.00\% & 82.19\%\\
   & 0.075 & 83.33\% & 366.24\% &  36.36\% & 82.92\%\\
   & 0.1 & 36.36\% & 776.81\% &  40.00\% & 83.80\%\\
  \hline
  \multirow{ 5}{*}{3} & 0.01 & -13.48\% & 108.84\% &  36.36\% & 49.87\%\\
   & 0.025 & 61.11\% & 116.60\% &  44.83\% & 52.01\%\\
   & 0.05 & 0.00\% & 159.91\% &  37.50\% & 61.71\%\\
   & 0.075 & 15.38\% & 201.90\% &  50.00\% & 59.12\%\\
   & 0.1 & -12.20\% & 122.82\% &  47.22\% & 58.73\%\\
  \hline
  \end{tabular}
  \label{greedyResult}
  }
\end{table}

As indicated by the table, using only greedy algorithm without choosing the initial points does not always find more counter-examples. However, the greedy algorithm greatly improves the numbers of counter-examples when choosing the initial inputs. Besides, compared with random mode, greedy is more dominant under the mode of choosing initial inputs. As indicated by Table \ref{result_mnist}, these results are also valid on MNIST networks. Hence, we can have the following answer for RQ2.
\vspace{-3mm}
\begin{framed}
\noindent \emph{Answer to RQ2:
The greedy algorithm is more effective and dominant when the initial input choosing is employed.
}\end{framed}

\subsubsection{Effectiveness of threshold generation method}
In principle, a DNN should has a specific difference threshold. Hence, we want to inspect the effectiveness of our threshold generation method. We compare the rates of finding counter-examples of our threshold generation method (denoted by \textsf{Minimum}) and that of the method uses the mean and the standard deviation, which generates the threshold to be the mean minus the standard deviation (denoted by \textsf{Average}). We randomly selected two DNNs from ASAC-Xu benchmark. For each one, we try to find 100 counter-examples. We generate the threshold from 100 randomly generated inputs, and $colNum$, \emph{i.e.}, collision number (\emph{c.f.} Algorithm \ref{Input}), is set to 1000.

%the following two threshold generation methods: 1) select the smallest distance from the randomly generated inputs, \emph{i.e.}, the one use

%Since the threshold of each neural network is different, a pre-processing is required to obtain the threshold we need before performing verification. We have designed two methods to get the threshold: Run $ReNum$ times and select the smallest distance as the threshold; Run $ReNum$ times, calculate the mean and standard deviation of the $ReNum$ distances, and set the threshold to the mean minus the standard deviation. We randomly selected two neural networks for testing, and each test was performed three times to reduce the mistake. For each test we tried to find 100 SAT inputs, and we set the $ReNum$ to 100 and the $ColNum$(collision num) to 1000.
%\vspace{-4mm}
\begin{table}[!ht]
  \centering
  \caption{Results of two methods for generating thresholds on ACAS-Xu networks.}
%  \vspace{-2mm}
  \begin{tabular}{cc|ccc}
  \hline
  Network & Method & \#Runs & Time(s) & Rate \\
  \hline
  1 & \textsf{Minimum} & 101 & 14620 & 98.36\% \\
  1 & \textsf{Average} & 107 & 18172 & 93.46\% \\
  \hline
  2 & \textsf{Minimum} & 107 & 21079 & 96.77\% \\
  2 & \textsf{Average} & 119 & 29870 & 85.71\% \\
  \hline
  \end{tabular}

  \label{selecting_thresholds}
\end{table}

Table \ref{selecting_thresholds} shows the results. Compared with \textsf{Average} method,  \textsf{Minimum} needs less time for finding 100 counter-examples, and achieves a higher rate of finding counter-examples, which explains why we use \textsf{Minimum} method. However, when the cost of generating a random input is high, \textsf{Minimum} method may not be better than \textsf{Average} method. In addition, there may exist other complicated threshold generation methods that are better than ours. However, since our method's cost is low and the rate of finding counter-examples is also pretty high, we believe we have achieved a balance.

\begin{framed}
\noindent \emph{Answer to RQ3:
Our threshold generation method is more effective than the average method.
}\end{framed}

%\begin{figure*}
%  \centering
%  \subfigure[$\delta$ = 0.01]{
%    \includegraphics[width=0.4\textwidth]{SAT_probability_0_01.png}
%    \label{delta0_01}
%  }
%  \quad
%  \subfigure[$\delta$ = 0.025]{
%    \includegraphics[width=0.4\textwidth]{SAT_probability_0_025.png}
%    \label{delta0_025}
%  }
%  \quad
%  \subfigure[$\delta$ = 0.05]{
%    \includegraphics[width=0.4\textwidth]{SAT_probability_0_05.png}
%    \label{delta0_05}
%  }
%  \quad
%  \subfigure[$\delta$ = 0.075]{
%    \includegraphics[width=0.4\textwidth]{SAT_probability_0_075.png}
%    \label{delta0_075}
%  }
%  \quad
%  \subfigure[$\delta$ = 0.1]{
%    \includegraphics[width=0.4\textwidth]{SAT_probability_0_1.png}
%    \label{delta0_1}
%  }
%  \caption{SAT probability of 4 methods in 5 neural networks under 5 $\delta$}
%  \label{SAT}
%\end{figure*}

\subsubsection{Effectiveness for DNN attacking methods}

We also inspect the effectiveness of our input selecting method for boosting the existing DNN attacking methods. Table \ref{attacking} shows the results. Column $eps$ shows the noise added to the input for each perturbation step. Column $epo$ shows how many perturbation steps are used for adversarial generation. Column Rate$_{\textsf{R}}$  shows the rate of successfully generating adversarial examples using random input selecting. Column Rate$_{\textsf{B}}$  shows that using our input selecting method.

%Effectiveness for DNN attacking methods, the Rate$_{ori}$ means the original successful attack rate, the Rate$_{imp}$ means the successful attack rate after use our method. the $epochs$ means maximum epochs to run, the $eps$ means the noise added to input per epoch
%  \vspace{-4mm}
\begin{table}[!ht]
  \centering
  \caption{Experimental results of boosting DNN attaching methods.}
%  \vspace{-2mm}
  \begin{tabular}{c|cc|cc}
  \hline
  Method & $eps$ & $epo$ & Rate$_{\textsf{R}}$ & Rate$_{\textsf{B}}$ \\
  \hline
  \multirow{ 6}{*}{FGSM} & 0.01 & 6 & 7.06\% & 58.50\%\\
  & 0.01 & 8 & 10.50\% & 63.67\%\\
  & 0.01 & 10 & 21.70\% & 68.70\%\\
  \cline{2-5}
  & 0.02 & 6 & 41.61\% & 65.66\%\\
  & 0.02 & 8 & 77.13\% & 92.12\%\\
  & 0.02 & 10 & 92.22\% & 93.34\%\\
  \hline
  \multirow{ 6}{*}{JSMA} & 1.0 & 10 & 16.50\% & 55.94\%\\
  & 1.0 & 20 & 43.13\% & 71.82\%\\
  & 1.0 & 30 & 70.88\% & 86.46\%\\
  \cline{2-5}
  & 2.0 & 10 & 15.45\% & 44.70\%\\
  & 2.0 & 20 & 51.74\% & 72.74\%\\
  & 2.0 & 30 & 66.20\% & 79.51\%\\
  \hline
  \multirow{ 6}{*}{CW} & 0.001 & 50 & 14.09\% & 48.77\%\\
  & 0.001 & 100 & 17.01\% & 64.43\%\\
  & 0.001 & 200 & 9.21\% & 57.36\%\\
  \cline{2-5}
  & 0.002 & 50 & 21.94\% & 66.10\%\\
  & 0.002 & 100 & 45.59\% & 73.19\%\\
  & 0.002 & 200 & 34.92\% & 69.23\%\\
  \hline
  \multirow{ 3}{*}{DeepFool} & none & 1 & 5.47\% & 56.10\%\\
  & none & 2 & 40.33\% & 91.73\%\\
  & none & 3 & 90.28\% & 93.35\%\\
  \hline
  \end{tabular}
  \label{attacking}
\end{table}
%  \vspace{-2mm}

As shown by the table, our input selecting method can improve the rate in all cases. In average, we can improve the success rate of generating adversarial examples by 3.2 times. Especially, when $eps$ and $epo$ are smaller, it is difficult for existing attacking methods to generate adversarial examples; while our input selecting method can achieve a better improvement in this situation.
%\vspace{-2mm}
\begin{framed}
\noindent \emph{Answer to RQ4:
Our input selecting method can improve the effectiveness of attacking methods. In average, our method can improve the success rate of generating adversarial examples by 3.2 times.
}\end{framed}

\subsection{Threats to Validity}

The threads to the validity of our work are mainly external. The external threats come from the following two aspects: 1) The selection of the verification tool and the benchmark DNN networks; however, Reluplex is not scalable to support large networks, which is the reason of why we use ASAC-Xu and MNIST networks; besides, if we randomly choose five ACAS-Xu networks, the main results do not change. 2) The selected attacking methods are limited.  We plan to apply our boosting method on more verification tools, benchmark networks and attacking methods in the future.

%!TEX root = main.tex
\section{Related Work}

Our work is related to the existing work of DNN attacking and defense, including verification, testing, adversarial example generation, \emph{etc}. %Next, the related work will be reviewed and compared with ours.

%verification
Existing DNN verification work can be divided into two categories: symbolic encoding based methods and abstraction based methods. Reluplex \cite{KatzBDJK17} is a representative work in the first category. Reluplex converts DNN's verification problems to an SMT solving problems. As far as we know, global robustness verification is not supported by Reluplex's current implementation. DeepSafe \cite{GopinathKPB18} proposes to cluster inputs to form input regions, and then use Reluplex to verify whether the regions are safe or not via local robustness verification. Similar to Reluplex, DLV \cite{HuangKWW17} also translates the verification problem of DNN to an SMT solving problem, and aims to verify robustness. DLV also needs an initial input to be given. AI$^2$ \cite{GehrMDTCV18} is a representative work in the second category. AI$^2$ \cite{GehrMDTCV18} uses abstract interpretation \cite{CousotC77} to verify the robustness of DNN. Using  different abstract domains, AI$^2$ abstracts the calculations and compositions in DNN, aiming to achieve a tradeoff between precision and scalability. ReluVal  \cite{WangPWYJ18} also uses interval abstract domain to verify DNNs to improve the scalability of robustness verification.  The approaches in the first category face scalability problem, especially on large-scale DNNs; however, they precisely encode the DNNs to verify, and the found counter-examples are real. On the other hand, the methods in the second category have a better scalability, but may suffer from false alarms. Compared with these verification work, our method is complementary, and can be combined with any verification work that supports local robustness verification.
% to improve the scalability.

%DNN testing (DeepXplore, DeepTest, DeepConcolic, DeepMutation, DeepGauge, TensorFuzz)
There are also many existing work of testing DNNs. Existing work includes testing criteria \cite{DeepXplore}\cite{MC/DC}\cite{DeepGauge}, automatic test generation \cite{DeepXplore}\cite{DeepTest}\cite{TensorFuzz}, test case measurement \cite{DeepMutation}, \emph{etc}. The basic idea is to apply the existing software testing techniques on DNN testing and propose new testing techniques specially for DNN testing. DeepXplore proposes \emph{neuron coverage} to measure DNN testing, and ultilizes differential testing \cite{XieTKM07} and gradient descent \cite{Kiwiel01} to automatically generate new inputs from existing inputs, aiming to improve neuron coverage and find adversarial examples. DeepTest focuses on the automatic testing of the DNNs in autonomous driving. DeepTest proposes a set of transformations specially designed for autonomous driving to generate new test cases. DeepTest considers that using the transformations can improve the neuron coverage of the DNN, and leverages metamorphic testing \cite{Chen1998Metamorphic} to address test oracle problem. DeepConcolic \cite{SunWRHKK18} provides a concolic testing \cite{GodefroidKS05}\cite{DBLP:conf/sigsoft/SenMA05} framework to automatically test DNN for improving different kinds of coverages and finding adversarial examples. DeepGauge \cite{DeepGauge} proposes a set of multi-granularity testing criteria for DNN to evaluate the testing adequacy; in \cite{MC/DC}, an MC/DC inspired coverage criterion is proposed. DeepMutation \cite{DeepMutation} is a DNN mutation testing technique  to evaluate the adequacy of DNN test cases. A set of mutation rules are proposed at different aspects of DNN, such as training code and model. TensorFuzz \cite{TensorFuzz} provides a coverage guided fuzzing framework for testing DNN to find errors.  How to combine our boosting method with the existing testing approaches is interesting and left to the future work.

%adversarial example related
To attack DNNs or improve the robustness of DNN, there exists many work for generating adversarial examples. \mbox{L-BFGS} \cite{SzegedyZSBEGF13} is the first method for generating adversarial examples. FGSM \cite{GoodfellowSS14} uses gradient update for generating adversarial examples. FGSM can generate an adversarial example from an input by just one update; hence, FGSM is pretty efficient. DeepFool \cite{Moosavi-Dezfooli16} utilizes Newton's iteration algorithm to generate a new input from a given one, aiming to change the output with least updates. Compared with FGSM, DeepFool has a higher probability of successfully generating adversarial examples but also a higher cost. JSMA \cite{PapernotMJFCS16} constructs a Jacobian matrix of given input, and modifies the input part that significantly influences output. CW \cite{Carlini017} provides an optimization-based attack method to defeat existing adversarial detection methods. Interestingly, there also exists a one-pixel attacking method \cite{onepixel} that modifies just one pixel of an image to generate an adversarial example. Our boosting method complements to the existing adversarial example generation methods. Our method can be used to select an input around which adversarial examples are more likely to exist. %; hence, our boosting method can improve both the effectiveness and the efficiency of adversarial example generation methods, which is already validated in our evaluation (\emph{c.f.}, Section \ref{exp}).
Beside attacking methods, there also exists work of detecting adversarial examples.
%, which can be used to filter suspicious adversarial examples to improve the safety and security of the DNN-based system. 
The authors in \cite{Xu0Q18} propose feature squeezing as a general framework to detect adversarial examples, and propose two squeezing instances for image classification. Besides, in \cite{DeepMutation}, mutation testing is used to detect adversarial examples, and the key observation is adversarial examples are more sensitive to perturbations.% than normal ones.

%!TEX root = main.tex
\section{Conclusion}

%Verifying DNN's safety is challenging. 
Existing DNN verification work suffers the scalability problem. In this paper, we focus on local robustness verification, and want to boost the finding of counter-examples during DNN verification. We propose a boosting method that generates the inputs around which there tend to exist counter-examples. Besides, we have proposed a greedy algorithm as a pre-analysis before verification to generate counter-examples. We have implemented our boosting method and carried out extensive experiments on representative benchmarks. The experimental results indicate the effectiveness of our  method. The future work mainly lies in more extensive evaluations on verification or attacking tools and other DNN benchmarks. 
%the following two aspects: 1) more extensive evaluation on different local robustness verification tools; 2) more extensive evaluation on more attacking methods and larger DNN benchmarks.% to validate the effectiveness further. 

\bibliographystyle{plain}
\bibliography{nuthin}

\end{document}